\RequirePackage{fix-cm}

\documentclass[twocolumn, natbib]{svjour3}

\smartqed  

\usepackage[ngerman,english]{babel}
\usepackage[utf8]{inputenc}

\usepackage{graphicx}
\usepackage[hidelinks]{hyperref}
\addto\extrasenglish{%
}

\usepackage{booktabs}
\usepackage{multirow}
\usepackage{colortbl}
\usepackage[figuresright]{rotating}
\usepackage{adjustbox}
\usepackage{placeins}  

\usepackage{xcolor}
\definecolor{semanticLabelUnknown}{rgb}{1.0,0.75,0.8}
\definecolor{semanticLabelRoadSurface}{rgb}{0.737,0.561,0.561}
\definecolor{semanticLabelGroundSurface}{rgb}{1.0,0.647,0.0}
\definecolor{semanticLabelCityFurniture}{rgb}{1.0,1.0,0.0}
\definecolor{semanticLabelVehicle}{rgb}{1.0,0.0,0.0}
\definecolor{semanticLabelPedestrian}{rgb}{0.188,0.329,0.588}
\definecolor{semanticLabelWallSurface}{rgb}{1.0,0.949,0.8}
\definecolor{semanticLabelRoofSurface}{rgb}{0.659,0.0,0.0}
\definecolor{semanticLabelDoor}{rgb}{0.776,0.349,0.067}
\definecolor{semanticLabelWindow}{rgb}{0.556,0.663,0.859}
\definecolor{semanticLabelBuildingInstallation}{rgb}{0.706,0.51,0.855}
\definecolor{semanticLabelSolitaryVegetationObject}{rgb}{0.329,0.514,0.208}
\definecolor{semanticLabelNoise}{rgb}{0.412,0.412,0.412}

\graphicspath{{./imgs/}}

\usepackage{amsmath,amssymb}
\usepackage{inputenc}
\usepackage[htt]{hyphenat}
\usepackage{comment}
\usepackage{graphicx}

\newcolumntype{L}[1]{>{\raggedright\let\newline\\\arraybackslash\hspace{0pt}}m{#1}}
\newcolumntype{C}[1]{>{\centering\let\newline\\\arraybackslash\hspace{0pt}}m{#1}}
\newcolumntype{R}[1]{>{\raggedleft\let\newline\\\arraybackslash\hspace{0pt}}m{#1}}

\usepackage{glossaries}
\usepackage{wasysym}

\newacronym{MLS}{MLS}{mobile laser scanning}
\newacronym{IoU}{IoU}{intersection over union}
\newacronym{mIoU}{mIoU}{mean intersection over union}
\newacronym{M3C2}{M3C2}{multi scale model to model cloud comparison}
\newacronym{C2C}{C2C}{cloud-to-cloud}
\newacronym{C2M}{C2M}{cloud-to-mesh}
\newacronym[plural=LODs,firstplural=Levels of Detail (LODs)]{LOD}{LOD}{Level of Detail}
\newacronym{ADAS}{ADAS}{Advanced Driver Assistant Systems}
\newacronym{AD}{AD}{Automated Driving}
\newacronym{DoF}{DoF}{Degrees of Freedom}
\newacronym{ML}{ML}{Machine Learning}
\newacronym{OGC}{OGC}{Open Geospatial Consortium}
\newacronym{GML}{GML}{Geographic Markup Language}
\newacronym{UAV}{UAV}{Unmanned Aerial Vehicle}
\newacronym{ALS}{ALS}{airborne laser scanning}
\newacronym{TLS}{TLS}{terrestrial laser scanning}
\newacronym{CAD}{CAD}{computer-aided design}
\newacronym{ASAM}{ASAM}{Association for Standardization of Automation and Measuring Systems}
\newacronym{FME}{FME}{Feature Manipulation Engine}
\newacronym{LiDAR}{LiDAR}{Light Detection and Ranging}
\newacronym{DoGSS-PCL}{DoGSS-PCL}{Domain Gap of Simulated Semantic Point Cloud}

\journalname{PFG - Journal of Photogrammetry, Remote Sensing and Geoinformation Science}

\begin{document}

\title{Mind the domain gap: Measuring the domain gap between real-world and synthetic point clouds for automated driving development}


\author{Nguyen Duc \and Yan-Ling Lai \and Patrick Madlindl \and Xinyuan Zhu \and Benedikt Schwab \and Olaf Wysocki \and Ludwig Hoegner \and Thomas H. Kolbe}


\institute{
N. Duc \and Y. Lai \and P. Madlindl \and X. Zhu \at TUM School of Computation, Information and Technology, Technical University of Munich, 80333 Munich, Germany \\
\and B. Schwab \and T.H. Kolbe \at Chair of Geoinformatics, Technical University of Munich, 80333 Munich, Germany \\
e-mail: benedikt.schwab@tum.de
\and
O. Wysocki \at Chair of Photogrammetry and Remote Sensing, Technical University of Munich, 80333 Munich, Germany \\
e-mail: olaf.wysocki@tum.de
\and
L. Hoegner \at 
Department of Geoinformatics, University of Applied Science Munich, 80333 Munich, Germany
}

\date{Received: date / Accepted: date}

\maketitle
\begin{abstract}

Owing to the typical long-tail data distribution issues, simulating domain-gap-free synthetic data is crucial in robotics, photogrammetry, and computer vision research.
The fundamental challenge pertains to credibly measuring the difference between real and simulated data.
Such a measure is vital for safety-critical applications, such as automated driving, where out-of-domain samples may impact a car's perception and cause fatal accidents.
Previous work has commonly focused on simulating data on one scene and analyzing performance on a different, real-world scene, hampering the disjoint analysis of domain gap coming from networks' deficiencies, class definitions, and object representation.
In this paper, we propose a novel approach to measuring the domain gap between the real world sensor observations and simulated data representing the same location, enabling comprehensive domain gap analysis.
To measure such a domain gap, we introduce a novel metric \emph{\acrshort{DoGSS-PCL}} and evaluation assessing the geometric and semantic quality of the simulated point cloud.
Our experiments corroborate that the introduced approach can be used to measure the domain gap.
The tests also reveal that synthetic semantic point clouds may be used for training deep neural networks, maintaining the performance at the 50/50 real-to-synthetic ratio.
We strongly believe that this work will facilitate research on credible data simulation and allow for at-scale deployment in automated driving testing and digital twinning.

\end{abstract}


\section{Introduction}
\label{sec:intro}

\sloppy 

\begin{figure*}[htb]
    \centering
    \includegraphics[width=1\linewidth]{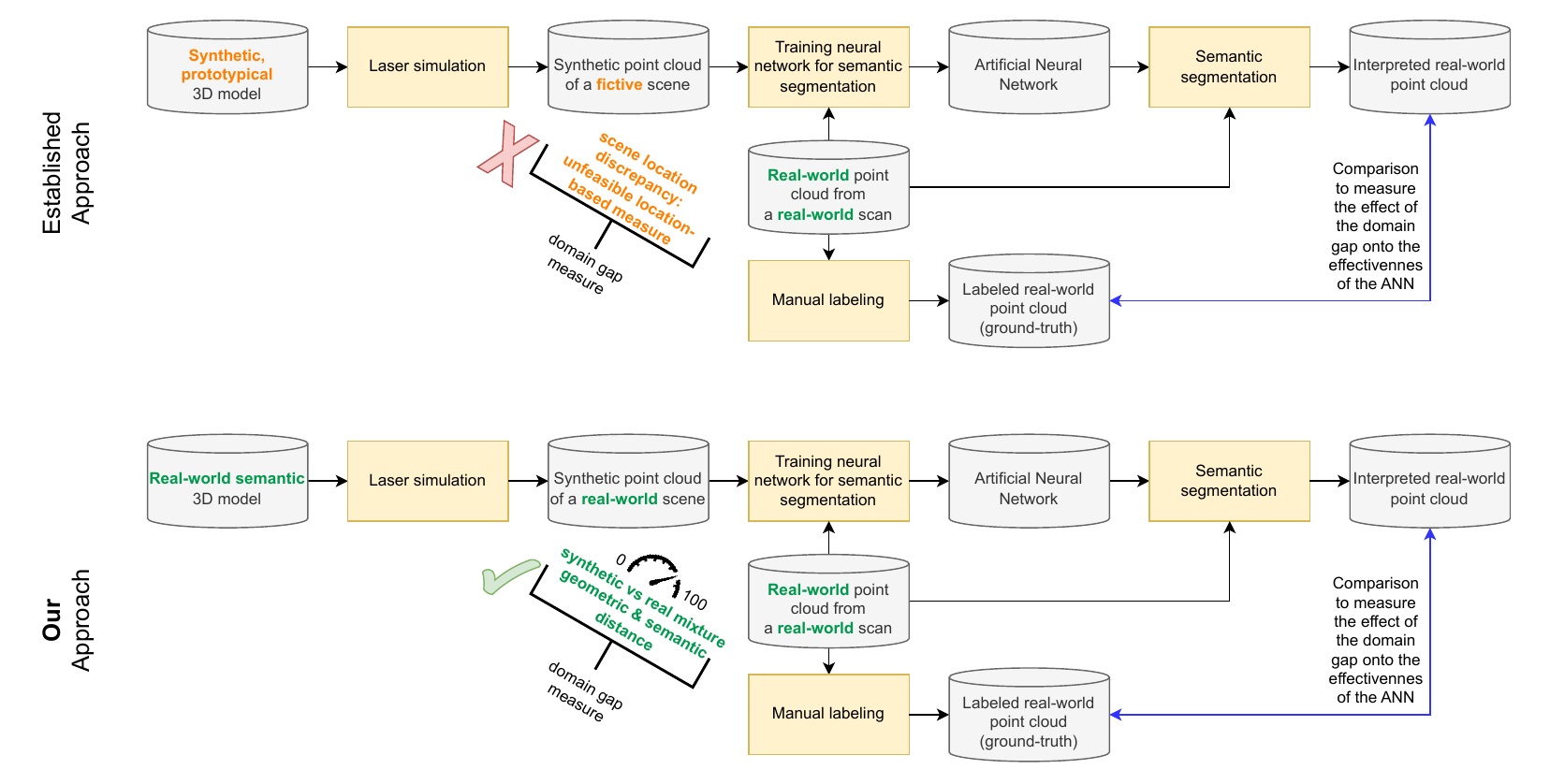}
    \caption{Unlike the established approaches measuring the domain gap on fictive 3D models and different, real-world locations (top branch);   
    we propose leveraging 3D models representing real-world cities and corresponding real-world point clouds for this purpose offering scene-homogeneous geometric and semantic domain gap measure (bottom branch). } 
    \label{fig:neat}
\end{figure*}
The scarcity of annotated 3D point cloud data poses a challenge in training robust models for various downstream tasks, such as urban semantic segmentation, object recognition, and 3D semantic reconstruction \citep{point_augment,hu2021towards,chen2022stpls3d,wysocki2023scan2lod3}.
It is especially apparent in the automated driving domain, where small data subsets are frequently used to verify the method's reliability, followed by the application on a broader scale.
To address the problem of data scarcity, researchers have considered various approaches, such as weakly-supervised  \citep{lin2022weakly,wang2024survey} or self-supervised \citep{zeng2024self,zhang2021self} point cloud understanding.

However, such approaches necessitate large validation datasets for development and testing, and their performance is still limited, e.g., reaching up to around 60\% accuracy for object detection \citep{zeng2024self}.
An alternative, simulator-oriented approach focuses on utilizing \gls{LiDAR} simulations leveraging virtual 3D environments for generating an arbitrary number of data samples with various distributions aiming at mitigating the long-tail class distribution issues \citep{carla_paris,synthcity}. 
While simulators have witnessed notable advancements recently, a significant domain gap persists between synthetic and real-world data \citep{mind_the_gap}. 

The term \textit{domain gap} frequently refers to the deviation between real-world and synthetic data, as one is collected from physical and the other from digital objects. 
The smaller this domain gap is, the larger the potential of synthetically generated data to serve as an alternative to real-world data \citep{synthcity,mind_the_gap}.
For example, the Paris-CARLA-3D dataset proposes mixing real and synthetic data for a more comprehensive class distribution \citep{carla_paris}.
This notion raises the questions: a) How does one determine that a simulated synthetic point cloud is closer to its real-world counterpart than another? b) To what extent can simulated data replace the real counterpart in real-world scenarios?

As we show in \autoref{fig:neat}, the domain gap is typically measured by simulating sensor observations in a generated virtual 3D scene, then applying such synthetic data to real-world data and testing the performance discrepancy.
However, this approach has several shortcomings: 
a) The virtual 3D scene is limited to a fictive scene created by a 3D designer and often procedural methods, not covering the complexity and perturbations of the real world;
b) Also, the class description in simulated and real data is frequently heterogeneous, impacting the validation analysis;
c) Ultimately, such an approach compounds sensor noise, virtual environment imperfections, and neural networks model's performance evaluation under one metric, hampering the comprehensive analysis of real versus simulated data gap.

In this work, we are the first to introduce a method to analyze the impact of synthetic data impact on real-world scenarios by a) Deploying an actual 3D model of the real urban layout, effectively elevating discrepancies coming from the fictive model; b) introducing harmonized classes between international modelling standards and target segmentation objectives; c) effectively, allowing to analyze the domain gap without compounding factors of fictive 3D models and heterogeneous training and validation classes, enabling comprehensive geometric and semantic analysis.
Moreover, we provide a framework for assessing the extent of complementary information gain stemming from simulated data to replace the costly real-world manual ground truth annotations.


Our contributions are summarized as follows:
\begin{itemize}
    \item Introducing a harmonized class list for real-world and simulated point clouds based on the international standards OGC CityGML 2.0 and ASAM OpenDRIVE
    \item Developing a pipeline to simulate \gls{LiDAR}  sensors  while preserving the semantic information according to the introduced point cloud class list
    \item Proposing a novel metric named \emph{\acrshort{DoGSS-PCL}} that measures the semantic and geometric quality of the synthetic point cloud
    \item Presenting a deterministic approach to evaluate the semantic and geometric domain gap of synthetic point clouds by applying the proposed metric
    \item Complementary analyses with a stochastic approach that measures the domain gap by comparing the performance of semantic segmentation neural networks on real-world and synthetic point clouds
\end{itemize}

\section{Related Work}
\label{sec:relatedWork}
\subsection{Driving Simulation and \gls{LiDAR} Sensor Models}

When developing and testing automated driving systems, submicroscopic driving simulators are utilized to simulate the vehicle's environment and its sensors.
Established driving simulators include IPG CarMaker \citep{ipgautomotivegmbhCarMaker2024}, Vires VTD \citep{vonneumann-coselVirtualTestDrive2014}, Vector DYNA4 \citep{vectorinformatikgmbhDYNA4VirtualTest2024}, and the open-source solution CARLA \citep{dosovitskiy2017carla}, which is based on the Unreal Game Engine.
These driving simulators have interfaces to the automated driving systems under test and are real-time capable.
A wide range of \gls{LiDAR} sensor models exists, which differ in terms of their modeling approach, sensor effects, and validation approach \citep{haiderDevelopmentHighFidelityAutomotive2022,hankeGenerationValidationVirtual2017,rosenbergerSequentialLidarSensor2020,zhaoMethodApplicationsLidar2021}.
In order to obtain point clouds as output, ray-tracing algorithms are commonly utilized to model the propagation of the laser beams in the virtual environment model.
For example, the \gls{LiDAR} sensor model supplied with the CARLA driving simulator also includes effects such as signal attenuation, drop-offs in the number of points, and loss due to external perturbations.

In addition to the sensor models for automated driving, there are also \gls{LiDAR} simulators, which stem from the remote sensing domain.
For example, HELIOS++ is a simulation framework to carry out virtual laser scanning campaigns for terrestrial, mobile, \gls{UAV}, and \gls{ALS} settings \citep{winiwarterVirtualLaserScanning2022}.
It supports the generation of full waveform data by sampling sub-rays inside the laser beam cone.
To model the interaction with matter, the material properties diffuse and specular scattering coefficients, as well as the reflectance for the object surfaces, are utilized, and the ray's reflected energy is computed with the bidirectional reflection distribution function according to Phong \citep{winiwarterVirtualLaserScanning2022,phongIlluminationComputerGenerated1975,jutziNormalizationLidarIntensity2009}.

\subsection{Semantic Modeling of the Urban Environment}

Most driving simulators support the import and export of OpenDRIVE-compliant road networks.
OpenDRIVE is a standard for describing the lane-level road networks for developing and validating advanced driver assistance systems as well as automated driving functions.
The standard is developed by the \gls{ASAM} and the current version, 1.8.0, was released in 2023 \citep{asamOpenDRIVEV1User2023}.
OpenDRIVE utilizes a linear referencing concept whereby all road objects are defined relative to the reference line of the respective road.
OpenDRIVE's data model is limited to representing road objects with rather abstract geometries.
For example, poles, trees, and traffic signals are commonly represented by an oriented 3D bounding box or cylinder geometry.
Road objects, such as buildings, controller boxes, and traffic islands, are typically represented by extruded polygonal outlines.
Such abstract geometric representations of road objects are sufficient for the execution of ground-truth, geometric, and stochastic sensor models \citep{magosiSurveyModellingAutomotive2022}.
However, phenomenological and physics-based sensor models require a comprehensive and gapless representation of the road spaces \citep{schwabRequirements2019}.
For such applications, the scene editors for driving simulators replace the coarse road object geometries with more detailed 3D assets and then export the graphical model alongside the OpenDRIVE dataset.

In order to represent, store, and exchange semantic models of entire cities for a broad spectrum of applications in 3D, the CityGML standard has become internationally established \citep{kolbeOGCCityGeography2021,biljeckiApplications3DCity2015,wysockiReviewingOpenData2024}.
The standard is issued by the \gls{OGC} and defines a conceptual data model as well as an encoding for the \gls{GML} \citep{kutznerOGCCityGeography2023}.
A city is hierarchically decomposed into its constituents and the city objects are modeled with respect to their geometric, semantic, and topological aspects as well as their appearance.
While buildings were primarily represented in CityGML on the basis of cadastral information in the past, the focus is currently expanding to encompass the semantic representation of road spaces.
CityGML supports multiple \glspl{LOD}, whereby building models in \acrshort{LOD}2 include generalized wall surfaces with roof shapes, and starting from \acrshort{LOD}3 the façades also encompass semantically differentiated windows, doors, and balconies.

\subsection{Transfer Learning on Point Cloud Data}
Traditional machine and deep learning methods require a large amount of training data, which is expensive to collect.
Especially in the context of semantic segmentation of 3D point clouds, the scarcity of well-annotated data for each sensor type has become a notable challenge \citep{are_we_hungry,mind_the_gap,xiao2022transfer}. 
The immediate application of a model trained on one sensor type to another is often infeasible owing to the different laser scanning patterns, point distribution, and point cloud size.
For instance, \cite{SegTrans} show semantic segmentation decreases its performance significantly when trained on \gls{TLS} and inferred on \gls{MLS} point cloud, i.e., 76.5\% to 32.0\% in their experiments.
Recent research endeavors have shifted towards integrating transfer learning techniques accounting for the sensor differences and showing promising results \citep{xiao2022transfer}, e.g., improving the results by up to 13\% in segmentation accuracy in the experiments of \cite{SegTrans}.

Yet, unlike in image-based transfer learning \citep{ros2016synthia}, there is a lack of large real-world point clouds to train a generic classifier that can be adapted for multiple scenarios \citep{wysockiTUMFACADE}.
Consequently, recent years have witnessed a surge in methods investigating the adoption of synthetic point clouds complementing real-world scenarios \citep{sim2real,xiao2022transfer,carla_paris}. 
However, it poses yet another challenge of not only accounting for the sensor type differences but also the domain gap between the simulated and real-world data. 
As reflected by experiments conducted by \cite{xiao2022transfer}, the performance for combining simulated and real data can oscillate in the range of 55-65\% accuracy depending on the scenario and benchmark dataset, underscoring the importance of further domain-gap investigations.

Although recently researchers have shown that weakly-supervised  \citep{lin2022weakly,wang2024survey} and self-supervised \citep{zeng2024self,zhang2021self} methods show promising results, there are still necessitating validation sets and their performance can be limited, e.g., can reach only up to 60\% when deployed for object detection, as shown in the review of \citep{zeng2024self}.

Another issue impacting transfer learning is the so-called negative transfer, which, according to \citep{wang2019characterizing} is formulated as \textit{transferring knowledge
from the source can have a negative impact on the target
learner}.
One of the key issues pertains to analyzing the negative transfer not according to the algorithm but only to the target data; in the case of synthetic data, following the standard way of training on synthetic and testing on real data (target). Another issue relates to disjoint distributions of two datasets. When labeled target data is limited, leveraging only marginal similarities is challenging, but with sufficient labeled data, a well-designed algorithm can minimize negative transfer by effectively aligning source and target distributions \citep{wang2019characterizing,ge2014handling}.
\begin{figure*}[htb]
    \centering
    \includegraphics[width=0.95\linewidth]{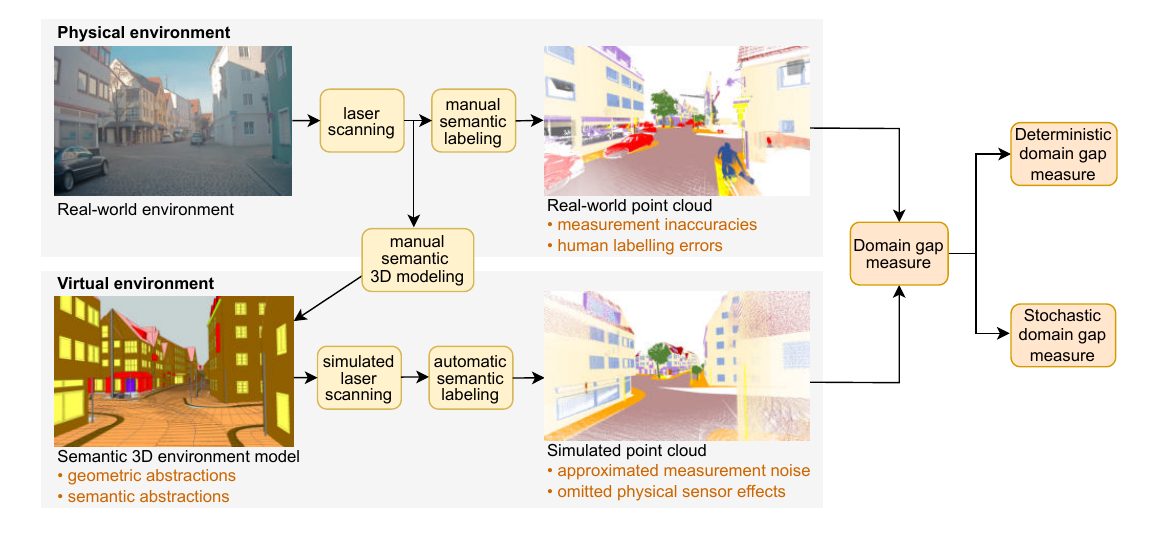}
    \caption{Overview of our domain gap measure workflow. We leverage the real-world point clouds and manually created semantic 3D urban models to identify deterministically (\autoref{sec:deterministicApproach}) and stochastically (\autoref{sec:stochasticApproach}) the point clouds domain gap. We also propose unified semantic labels for both 3D-model-simulated and real-world point clouds in accordance with international 3D modeling standards (\autoref{subsec:semanticLabelsMapping}).} 
    \label{fig:architecture}
\end{figure*}
\begin{figure*}[htb]
    \centering
    \includegraphics[width=0.95\linewidth]{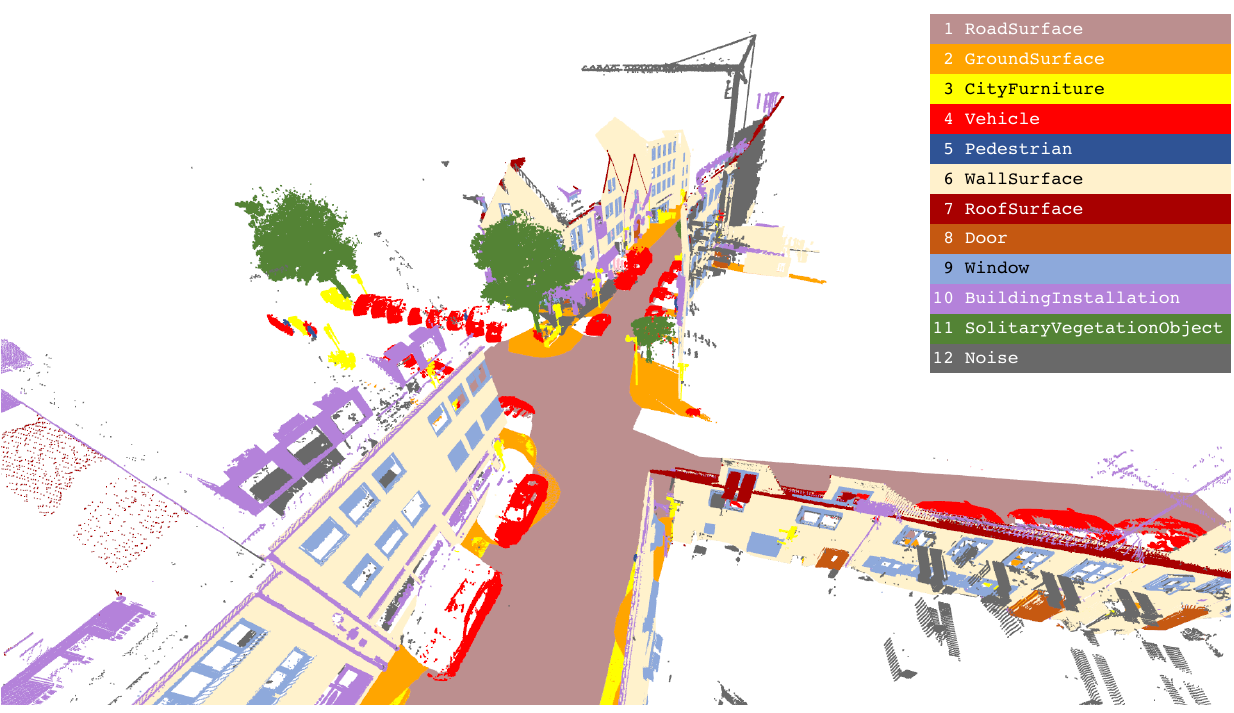}
    \caption{Real-world point cloud, which was manually labeled according to the class list of \autoref{tab:class-list}.} 
    \label{fig:detailedLabledPointCloud}
\end{figure*}
\subsection{Domain Gap between Synthetic and Real-World Point Cloud}
In recent years, different efforts to evaluate the domain gap for point cloud data have been undertaken \citep{carla_paris,carla_adaSplats,huch2023quantifying}.
As point cloud semantic segmentation is one of the fundamental challenges, the domain gap is frequently investigated by comparing the simulated to real-world data segmentation performance \citep{xiao2022transfer}. 
The typical techniques range from domain adaptation, domain randomization, transfer learning, feature space alignment, generative methods, and hybrid training \citep{xiao2022transfer,huch2023quantifying,guan2021domain}.

Simultaneously, the virtual testbed fidelity is regarded as crucial for simulating point clouds reflecting reality \citep{yue2018lidar}. 
For instance, \cite{carla_paris} have shown a method to create synthetic data based on \gls{CAD} 3D models and the CARLA simulator.
They also propose a strategy for reconstructing the environment from authentic point clouds using splats \citep{carla_adaSplats}, aiming to create imperfect synthetic scans that resemble real-world point clouds. 
Still, however, the difference in semantic segmentation performance is significant; when trained on the well-established KPConv \citep{thomas2019kpconv}, the accuracy drops by up to 38\% on their Paris-CARLA-3D dataset when compared to the real data segmentation performance \citep{carla_paris}. 

One of the flaws of such virtual worlds is that they do not represent any actual city but rather an imaginative city scene created by a 3D designer \citep{dosovitskiy2017carla}.
Also, \gls{CAD}-generated scenes have only limited semantic information, which hinders its immediate extraction for training deep learning models, in contrast to CityGML-based models describing hierarchical semantics of each surface of a 3D model \citep{Kolbe2021}.

To the best of our knowledge, no publication investigates both the semantic and geometric simulated point cloud quality deterministically and stochastically using semantic 3D city models representing actual cities.

\section{Methodology}
\label{sec:methodology}
As shown in \autoref{fig:architecture}, we utilize real-world and synthetic point clouds simulated in the corresponding semantic 3D environment model to evaluate the domain gap.
To achieve this, we introduce a set of semantic labels in \autoref{subsec:semanticLabelsMapping} that is derived from the semantic modeling standards and is consistently applied to both the real-world and the synthetic point clouds.
To systematically evaluate the domain gap, we introduce a deterministic approach in \autoref{sec:deterministicApproach} and a stochastic approach in \autoref{sec:stochasticApproach}.


\subsection{Semantic Point Cloud Classes}
\label{subsec:semanticLabelsMapping}
A consistent class list is required to ensure that the real-world point clouds are comparable with their simulated counterparts.
Concerning our research objectives, the class list must fulfill the following requirements:
\begin{itemize}
    \item 
    The point clouds and the class list shall be tailored to train neural networks for semantic segmentation to analyze the domain gap stochastically. A typical problem that may influence the neural network's performance is data imbalance or skewed class proportions, often referred to as the long-tail distribution issue \citep{dai2017scannet}. Therefore, ideally, the number of objects in each class shall be approximately the same.
    \item  While we desire the class list to be sufficiently detailed, it shall not be overly redundant. Instead, it shall utilize a relatively limited number of classes describing all the critical object types that might be encountered in an urban environment. Therefore, we design our classes based on the internationally renowned urban modeling standards developed by researchers and practitioners, such as CityGML \citep{kolbeOGCCityGeography2021,grogerOGCCityGeography2012} and OpenDRIVE \citep{asamOpenDRIVEV1User2023}, as well as related works \citep{wysockiTUMFACADE,archDatasetPaper}. 
\end{itemize}
Regarding the requirements mentioned above, we present the description of our introduced 12 classes in \autoref{tab:identifiedClasses}.
Furthermore, we present the relation of the classes to the international OpenDRIVE 1.4 and CityGML 2.0 standards in \autoref{tab:class-list}.
\begin{table}[htb]
    \caption{Description of our introduced 12 road space classes.}
    \label{tab:identifiedClasses}
    \footnotesize
    \centering
    \newcommand{\colorBox}[1]{%
        \textcolor{#1}{$\blacksquare$}
    }
    \begin{tabular}{@{} L{.08\linewidth} L{.85\linewidth} @{}}
        \toprule
        ID & Class/ description \\
        \midrule
        1 \colorBox{semanticLabelRoadSurface} & \texttt{RoadSurface} \\
        & Describes any vehicle-allowed surfaces, w/o sidewalks and other road installations\\
        2 \colorBox{semanticLabelGroundSurface} &\texttt{GroundSurface} \\
        & Pedestrian-allowed surfaces, w/o roads\\
        3 \colorBox{semanticLabelCityFurniture} &\texttt{CityFurniture} \\
        & Any vertical urban installation, w/o building-attached objects\\
        4 \colorBox{semanticLabelVehicle} & \texttt{Vehicle} \\
        & Any vehicle, parked or moving\\
        5 \colorBox{semanticLabelPedestrian} & \texttt{Pedestrian} \\
        & Any person, standing or moving\\
        6 \colorBox{semanticLabelWallSurface} & \texttt{WallSurface} \\
        & Vertical and planar building parts, w/o roofs, installations, and facade elements\\
        7 \colorBox{semanticLabelRoofSurface} & \texttt{RoofSurface} \\
        & Building parts forming roof structures\\
        8 \colorBox{semanticLabelDoor} & \texttt{Door} \\
        & Any opening allowing entering objects, w/ gates\\
        9 \colorBox{semanticLabelWindow} & \texttt{Window} \\
        & Any opening and its outer blinds, w/o entries\\
        10 \colorBox{semanticLabelBuildingInstallation} & \texttt{BuildingInstallation} \\
        & Any building-attached installation\\
        11 \colorBox{semanticLabelSolitaryVegetationObject} & \texttt{SolitaryVegetationObject} \\
        & Any vegetation, w/ tree trunks and branches\\
        12 \colorBox{semanticLabelNoise} & \texttt{Noise} \\
        & Noisy points and any other non-annotated element\\
        \bottomrule
    \end{tabular}
\end{table}
\begin{table*}[htb]
    \renewcommand{\arraystretch}{1.05}
    \newcommand{\hlineafter}{%
        \hline
        \noalign{\vskip 0.5mm}
    }
    \newcommand{\colorBox}[1]{%
        \color{#1}$\blacksquare$
    }
    \footnotesize
    \centering
    \caption{Correspondences between the classes of OpenDRIVE 1.4 and CityGML 2.0, and their mapping to the introduced point cloud classes. The colors are used in the remaining illustrations.}
    \label{tab:class-list}
    \begin{tabular}{@{} l l r l @{}}
        \toprule
        \multicolumn{2}{c}{Semantic 3D environment model} & \multicolumn{2}{c}{Point clouds (real-world and simulated)} \\
        \cmidrule(r){1-2} \cmidrule(l){3-4}
        OpenDRIVE 1.4 & CityGML 2.0 & ID & Class \\
        \midrule
        \noalign{\global\arrayrulewidth=0.2mm}\arrayrulecolor{gray!40}
        \texttt{LaneSectionLRLane} (type: driving) & \texttt{TrafficArea} (function: 1) & 1\,\colorBox{semanticLabelRoadSurface} & \texttt{RoadSurface} \\
        \texttt{RoadObject} (type: barrier, name: raisedMedian) & \texttt{AuxiliaryTrafficArea} & & \\
        \texttt{RoadObject} (type: barrier, name: trafficIsland) & \texttt{AuxiliaryTrafficArea} & \\
        \texttt{RoadObject} (type: roadMark) & \texttt{AuxiliaryTrafficArea} & \\
        \hlineafter
        \texttt{LaneSectionLRLane} (type: sidewalk) & \texttt{TrafficArea} (function: 2) & 2 \colorBox{semanticLabelGroundSurface} & \texttt{GroundSurface} \\
        \texttt{LaneSectionLRLane} (type: border) & \texttt{AuxiliaryTrafficArea} \\
        \texttt{LaneSectionLRLane} (type: none, material: grass) & \texttt{AuxiliaryTrafficArea} \\
         & \texttt{OuterFloorSurface} \\
        \hlineafter
        \texttt{Signal} (name: trafficLight) & \texttt{CityFurniture} & 3 \colorBox{semanticLabelCityFurniture} & \texttt{CityFurniture} \\
        \texttt{Signal} (name: traffic signs) & \texttt{CityFurniture} \\
        \texttt{RoadObject} (type: pole, name: streetLamp) & \texttt{CityFurniture} \\
        \texttt{RoadObject} (type: pole, name: trafficLight) & \texttt{CityFurniture} \\
        \texttt{RoadObject} (type: pole, name: trafficSign) & \texttt{CityFurniture} \\
        \texttt{RoadObject} (type: barrier, name: fence) & \texttt{CityFurniture} \\
        \texttt{RoadObject} (type: obstacle, name: controllerBox) & \texttt{CityFurniture} \\
        \texttt{RoadObject} (type: obstacle, name: bench) & \texttt{CityFurniture} \\
        \texttt{RoadObject} (type: barrier, name: wall) & \texttt{CityFurniture} & \\
        \hlineafter
        -- & -- & 4 \colorBox{semanticLabelVehicle} & \texttt{Vehicle} \\
        \hlineafter
        -- & -- & 5 \colorBox{semanticLabelPedestrian} & \texttt{Pedestrian} \\
        \hlineafter
        \texttt{RoadObject} (type: building, surf.\ orientation: side) & \texttt{WallSurface} & 6 \colorBox{semanticLabelWallSurface} & \texttt{WallSurface} \\
        \hlineafter
        \texttt{RoadObject} (type: building, surf.\ orientation: top) & \texttt{RoofSurface} & 7 \colorBox{semanticLabelRoofSurface} & \texttt{RoofSurface} \\
        \hlineafter
        -- & \texttt{Door} & 8 \colorBox{semanticLabelDoor} & \texttt{Door} \\
        \hlineafter
        -- & \texttt{Window} & 9 \colorBox{semanticLabelWindow} & \texttt{Window} \\
        \hlineafter
        -- & \texttt{BuildingInstallation} & 10 \colorBox{semanticLabelBuildingInstallation} & \texttt{BuildingInstallation} \\
        -- & \texttt{OuterCeilingSurface} & \\
        \hlineafter
        \texttt{RoadObject} (type: tree) & \texttt{SolitaryVegetationObject} & 11 \colorBox{semanticLabelSolitaryVegetationObject} & \texttt{SolitaryVegetationObject} \\
        \texttt{RoadObject} (type: vegetation) & \texttt{SolitaryVegetationObject} \\
        \hlineafter
        -- & -- & 12 \colorBox{semanticLabelNoise} & \texttt{Noise} \\
        \arrayrulecolor{black}
        \bottomrule
    \end{tabular}
\end{table*}

\subsection{Harmonizing the Virtual Environment}
\label{subsec:SemModelGeneration}
\begin{figure}[htb]
    \centering
    \includegraphics[width=\linewidth]{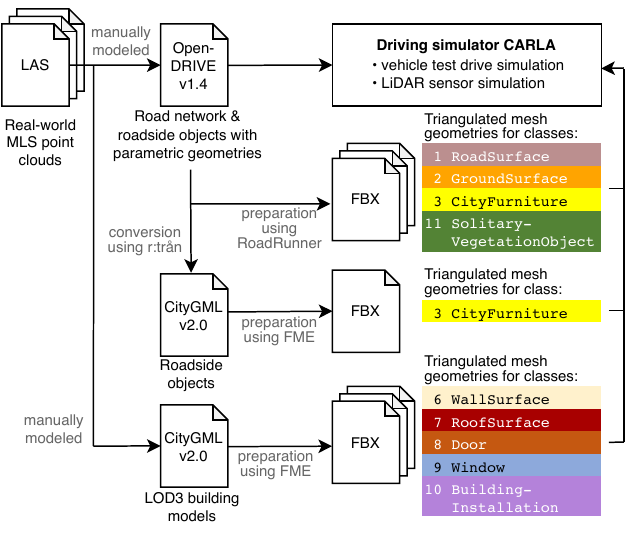}
    \caption{Developed model processing chain that enables a \gls{LiDAR} scanning simulation with automatically assigned labels according to our proposed class list. The road network standard OpenDRIVE and the semantic 3D city model standard CityGML serve as the basis for deriving application-specific mesh geometries.} 
    \label{fig:modelProcessing}
\end{figure}

In order to simulate a \gls{LiDAR} sensor in the virtual environment and to obtain the semantic labels automatically, the semantic environment models need to be prepared accordingly for the simulator.
As the domain gap is specifically investigated for the development and testing of automated driving, the submicroscopic open-source driving simulator CARLA is employed for the sensor simulation.
CARLA supports the simultaneous simulation of further exteroceptive sensors, such as RADAR and camera, which are required for testing the perception system and for further potential downstream domain gap analyses.
To derive an environment model representation for CARLA while preserving the relevant semantic labels of the objects, we developed a model preparation and processing chain shown in \autoref{fig:modelProcessing}.

CARLA imports the road network description from OpenDRIVE datasets, which enables the simulated vehicle agents to plan routes and driving trajectories at the lane level.
In order to simulate the \gls{LiDAR} sensor in CARLA with automatic labeling according to our proposed class list, a triangulated mesh geometry representation of the environment is required, whereby the semantic labels must be retained.

As depicted in \autoref{fig:modelProcessing}, we utilized three processing paths to derive triangulated mesh geometries from the semantic 3D environment models.
First, we used the 3D scene editor \emph{RoadRunner} \citep{roadrunner} to generate mesh geometry representations for the road surfaces and ground surfaces from the OpenDRIVE dataset.
This also includes traffic lights and signs, which are classified under \texttt{CityFurniture}, and trees under the class \texttt{SolitaryVegetationObject}.
Since \emph{RoadRunner} does not support the full range of road objects that are represented in OpenDRIVE, we use the open-source tool \emph{r:trån}\footnote{Website: \url{https://rtron.io}} to convert the OpenDRIVE dataset to CityGML 2.0 \citep{su12093799}.
This second path includes fence and road-boundary wall objects, whereby the geometries are triangulated and combined using the tool \gls{FME}.
In the third path, the \acrshort{LOD}3 building models are utilized, which have been modeled manually on the basis of the real-world point clouds and are already available as CityGML 2.0 datasets.
Thereafter, the object geometries are also triangulated using the tool \gls{FME} and the mesh geometries are stored separately according to our proposed class list.

Since the OpenDRIVE and CityGML datasets are georeferenced, the geometries were translated into a local coordinate reference system for the driving simulator.
As a result, we have a simulation environment derived from the semantic 3D model that reflects the physical environment with relative accuracy in the low centimeter range, as shown in the overview \autoref{fig:architecture}.

\subsection{Simulating Laser Scanning}
\label{subsec:simulatingLaser}
\begin{figure}[htb]
    \centering
    \includegraphics[width=1.0\linewidth]{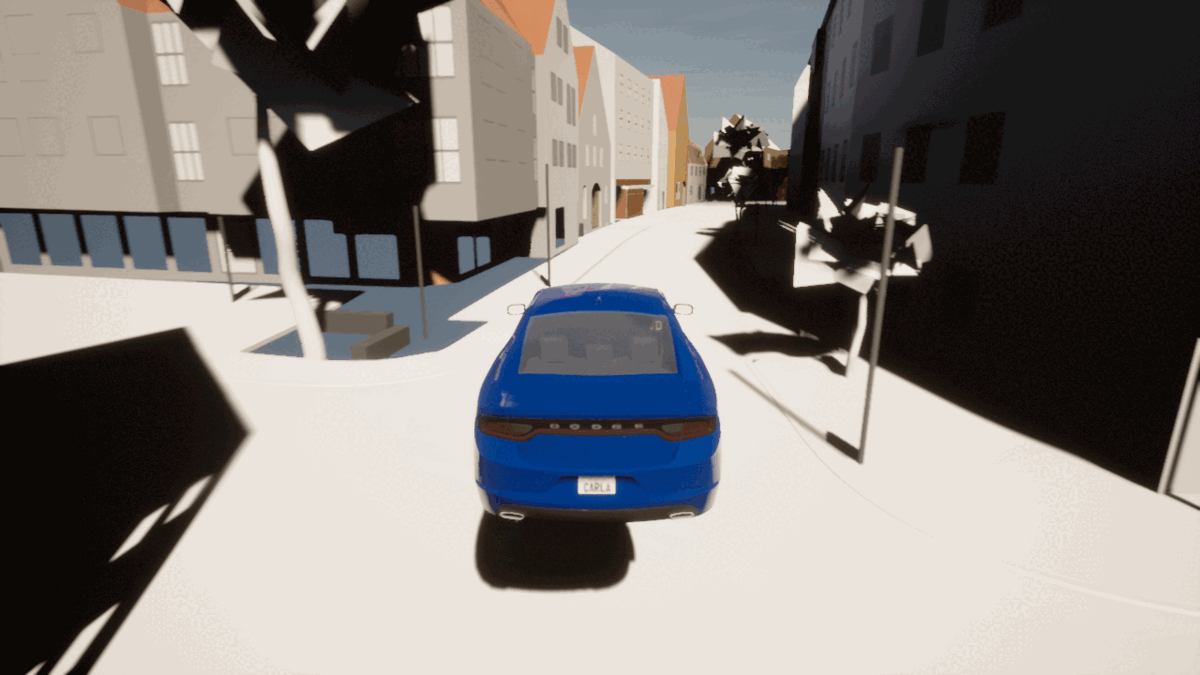}
    \includegraphics[width=1.0\linewidth]{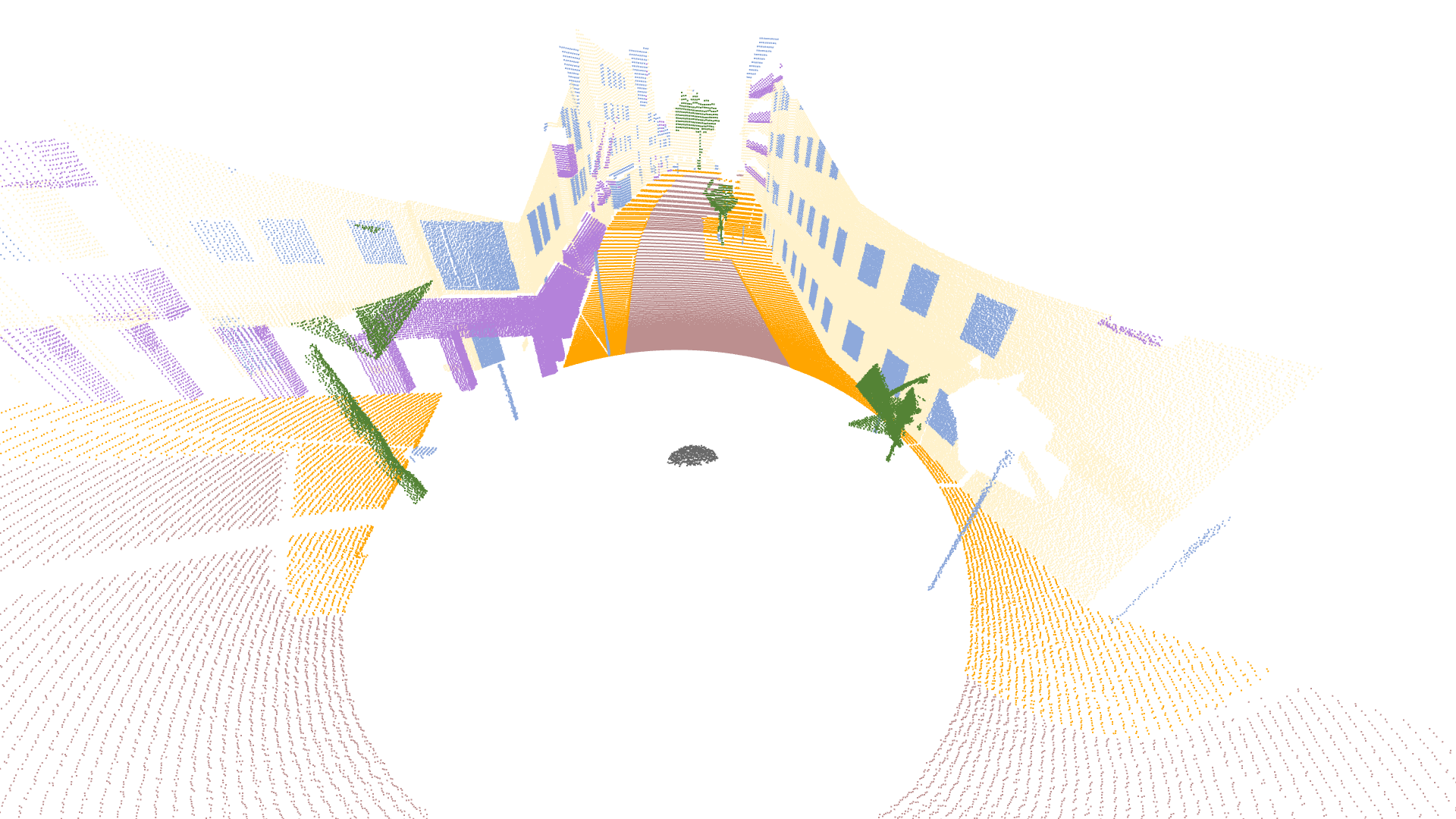}
    \caption{Simulated test drive in the virtual environment with the automatically labeled point cloud after applying noise in the post-processing.
    The virtual vehicle was elevated to match the sensor positions of the real-world surveying vehicle.}
    \label{fig:carla_simulator}
\end{figure}

After importing the virtual environment into the submicroscopic driving simulator, we set up a virtual vehicle with \gls{LiDAR} sensors to collect the point cloud data.
CARLA supports the configuration of the vehicle's sensor suite and environmental conditions to reproduce a driving scenario.
Since the exact sensor configuration was not disclosed by the company that conducted the real-world \gls{MLS} campaign, we have approximated the sensor positions and configuration parameters as closely as possible based on \citep{mappingSolutions2023Moses,grafeHighPrecisionKinematic2007}.
\autoref{fig:carla_simulator} shows the simulated vehicle test drive and the corresponding point cloud obtained from this timestamp.

As CARLA's semantic LiDAR sensor model lacks built-in noise support, we introduce noise during post-processing by applying a Gaussian distribution to the distance measurements along the raycasted vector.
We assume that the distance measurement error is normally distributed with a standard deviation of $\rho = 2$~cm, following \cite{carla_noise}.
\begin{figure}[htb]
    \centering
    \includegraphics[width=1.0\linewidth]{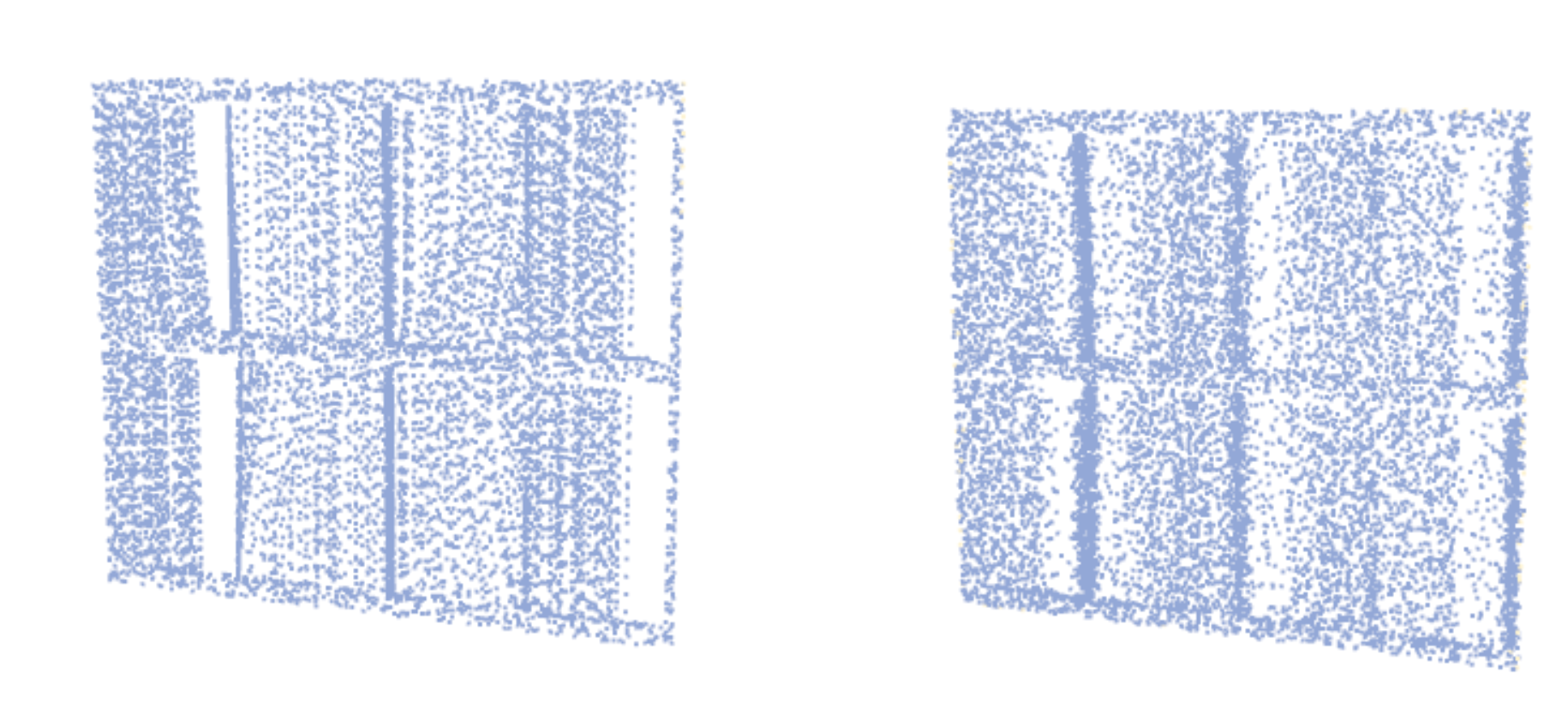}
    \caption{Scan of a window with curtains in the interior before (left) and after (right) adding noise.}
    \label{fig:noise_example}
\end{figure}
\autoref{fig:noise_example} shows the simulated point cloud of a window object with curtains before and after applying normal-distributed noise to the distance measurements. 

\subsection{Measuring Domain Gap Between Real-World Data and Synthetic Data}
\label{sec:MeasuringDomainGap}
We propose a deterministic and stochastic approach to investigate the concept of domain gap.
The deterministic approach involves computing distances and designing a custom metric named \emph{\gls{DoGSS-PCL}} to quantify how well the synthetic point clouds match the real-world ones. 
The stochastic approach involves utilizing synthetic and real-world point clouds in a common deep-learning-based task: 3D semantic segmentation. 
We focus on the 3D point cloud semantic segmentation while assessing its inference performance, leveraging well-established metrics in the semantic segmentation field.
Here, the real-world point cloud serves as the ground-truth dataset.

\subsubsection{Deterministic Approach} 
\label{sec:deterministicApproach}
To compare the real-world point cloud $\mathcal{R}$ with the synthetic point cloud $\mathcal{S}$ deterministically, we compute
\begin{itemize}
    \item[$(1)$] \textbf{Semantic-Based Point Cloud Distance (SBPCD)} $d(\mathcal{R},\mathcal{S})$ to measure between the point clouds $\mathcal{R}$ and $\mathcal{S}$
    \item[$(2)$]  \textbf{Domain Gap of Simulated Semantic Point Cloud (\acrshort{DoGSS-PCL} Metric}) $m_{\acrshort{DoGSS-PCL}}(\mathcal{R},\mathcal{S})$ that allows to compare and benchmark synthetic point clouds $\mathcal{S}_i$ given one corresponding real-world point cloud $\mathcal{R}$ (2).
\end{itemize}

\paragraph{\textbf{(1) Semantic-Based Point Cloud Distance}} We choose to compute and weigh two distance measures. 
\\
(a) First, a \acrlong{C2C} distance $d_{C2C}(\mathcal{R}, \mathcal{S})$ between the two point clouds $\mathcal{R}, \mathcal{S}$ \citep{C2C,c2cFormula}:
\begin{equation}
d_{C2C}(\mathcal{R}, \mathcal{S}) = \max_{a \in \mathcal{R}_p} \left\{ \min_{b \in \mathcal{S}_p} \{ d(a_\mathcal{R}, b_\mathcal{S}) \} \right\},
\end{equation}
where \(a_\mathcal{R}\) are the points of set \(\mathcal{R}_p\); \(b_\mathcal{S}\) are the points of set \(\mathcal{S}_p\); \(d(a_\mathcal{R}, b_\mathcal{S})\) is the Euclidean distance between \(a_\mathcal{R}\) and \(b_\mathcal{S}\).

(b) Secondly, making use of semantic information present in the point clouds, we compute a mean class-wise \acrshort{M3C2} distance $d_{MM3C2}(\mathcal{R}, \mathcal{S})$ \citep{M3C2}. Let $C$ be the number of classes in both $\mathcal{R}$ and $\mathcal{S}$. Then we split the point clouds into class-wise point clouds, such that $\mathcal{R} = \bigcup\limits_{c=1}^{C} \mathcal{R}^{(c)}$ and $\mathcal{S} = \bigcup\limits_{c=1}^{C} \mathcal{S}^{(c)}$. On each pair $(\mathcal{R}^{(c)}, \mathcal{S}^{(c)})$, we perform \acrshort{M3C2} and take the median of inlier M3C2 distances: $\mathbf{\widetilde{d}_{M3C2}(\mathcal{R}^{(c)}, \mathcal{S}^{(c)})}$. Outlier points, i.e., points where M3C2 has found no matches, are ignored. 
We then average the absolute values of the medians with custom class weights to enable balancing the importance of each class in the evaluation $\{w^{(c)}\}_{c=1}^{C}$  where $\sum_{c = 1}^{C} w^{(c)} = 1$:\\
\begin{equation}
\label{eq: mean_m3c2}
    d_{MM3C2}(\mathcal{R}, \mathcal{S}) = \sum_{c = 1}^{C} w^{(c)} \cdot \mathbf{\widetilde{d}_{M3C2}(\mathcal{R}^{(c)}, \mathcal{S}^{(c)})}
\end{equation}
\\
The distance $d_{\mathcal{R}, \mathcal{S}}$ is a weighted mean between the absolute value of $d_{C2C}(\mathcal{R}, \mathcal{S})$ and $d_{MM3C2}(\mathcal{R}, \mathcal{S})$:
\\
\begin{equation}
\label{eq:final_distance}
d(\mathcal{R}, \mathcal{S}) = \lambda_1 d_{MM3C2}(\mathcal{R}, \mathcal{S}) + \lambda_2 d_{C2C}(\mathcal{R}, \mathcal{S})
\end{equation}
\\
The distance has the same physical unit as point clouds $\mathcal{R}$ and $\mathcal{S}$ and represents an absolute Euclidean deviation between the clouds in the world frame (L2 distance).
\paragraph{\textbf{(2) \acrshort{DoGSS-PCL} Metric}} We combine $d_{C2C}(\mathcal{R}, \mathcal{S})$ and $d_{MM3C2}(\mathcal{R}, \mathcal{S})$ with a semantic measure, the weighted \gls{mIoU}, to form the metric $m_{\mathcal{R}}(\mathcal{S})$, while following the standard IoU formulation (analogical to Eq. 6):
\begin{equation}
    \label{eq: MIoU}
    mIoU(\mathcal{R}, \mathcal{S}) = \sum_{c = 1}^{C} w^{(c)} \cdot IoU(\mathcal{R}^{(c)}, \mathcal{S}^{(c)}).
\end{equation} 
We normalize with the min-max normalization the result to the interval $[0,1]$, where $0$ indicates the best possible match, and $1$ translates to the worst match. $mIoU(\mathcal{R}, \mathcal{S})$ behaves inversely in the sense that good values are close to $1$ and poor values close to $0$. We thus use an \acrshort{mIoU} factor $f_{mIoU}(\mathcal{R}, \mathcal{S}) = \frac{1}{mIoU(\mathcal{R}, \mathcal{S}) + \varepsilon}$, where $\varepsilon >0$ and should be chosen empirically and small to improve the numerical stability. 
Let the growth function be bounded by exponential upper-bound \citep{mohri2018foundations} controlled by empirically set $\alpha$ distance weight, defining the metric as (subtracted by one to normalize and obtain reverse score):

\begin{equation}
\label{eq:metric}
\begin{aligned}
\scriptsize 
m_{\text{DoGSS-PCL}}(\mathcal{R},\mathcal{S}) &= 1 - \exp\Bigg[ \alpha( d(\mathcal{R},\mathcal{S}) + \lambda_3 f_{mIoU}) \Bigg]
\end{aligned}
\end{equation}
The \gls{DoGSS-PCL} metric enables measuring different synthetic point clouds $\mathcal{S}_1, ..., \mathcal{S}_n$ and comparing their $m_{\text{DoGSS-PCL}}(\mathcal{R},\mathcal{S})$ scores.

The  \( \lambda_1, \lambda_2, \lambda_3 \) express weights assigned to the specific characteristic, which balance:  \( \lambda_1 \) importance for the class-wise change;  \( \lambda_2 \) cloud-to-cloud L2 distance; \( \lambda_3 \) semantic discrepancies.
Let  \( \lambda_1 + \lambda_2 + \lambda_3  \in [0,1] \) where \( \lambda_1 + \lambda_2 + \lambda_3  = 1 \) and shall follow \( \lambda_1 > \lambda_2  > \lambda_3 \) allowing for balanced weighting of the selected parameters and features.
Also, the weights should be chosen concerning balancing distance scores and importance of $\alpha( d(\mathcal{R},\mathcal{S})$.   

\subsubsection{Stochastic Approach}
\label{sec:stochasticApproach}
While deterministic measurement quantitatively assesses the similarity between real-world and synthetic point clouds, it does not give full intricacies for the frequently sought-after downstream scene understanding tasks. 
One of the prominent examples of data-scarce tasks is point cloud semantic segmentation \citep{wysocki2023scan2lod3}.
We leverage the established deep-learning-based 3D point cloud semantic segmentation models as benchmarks and utilize the established evaluation metrics to quantify and compare the network training performance. 
We propose mixed setups of point clouds with different proportions of real-world and synthetic point clouds that gradually amplify the influence of synthetic point clouds on the downstream segmentation task. 
These setups with the same quantity but increasing proportions of synthetic point clouds manifest the trade-off between model performance and the usage of synthetic point clouds.

We evaluate the model with class-wise $IoU^{(c)}$, which \gls{IoU} of class $c$ is calculated as follows:
\begin{equation}
    \label{eq: IoU-nn}
    IoU^{(c)} = \frac{TP^{(c)}}{TP^{(c)}+FP^{(c)}+FN^{(c)}}.
\end{equation}
where $TP^{(c)}$, $FP^{(c)}$, and $FN^{(c)}$ stand for true positive, false positive, and false negative, respectively. 
We use \gls{mIoU} to evaluate the learning and classification quality of the models on each training set, which is calculated by taking the average of class-wise $IoU^{(c)}$.
We assess the influence of the proportion of synthetic point clouds on the class-wise $IoU^{(c)}$ performance.
We evaluate the Pearson correlation coefficient between the class-wise $IoU^{(c)}$ and the synthetic point cloud proportion. 
Let $i$ be the index of the training set that consists of $p_{i}$ percent of synthetic data, $I$ be the set that contains all of the indexes of the training sets, and $P$ be the set contains all of $p_{i}$. 
The index $IoU^{(c)}_{i}$ represents the $IoU^{(c)}$ performance of set $i$ and $IoU_{c}$ represents the set contains all of $IoU^{(c)}_{i}$. 
We calculate the correlation coefficient of class $c$ as follows:
\begin{equation}
    corr^{(c)} = \frac{cov(P, IoU_{c})}{\sigma_{P} \sigma_{IoU^{(c)}}}.
\end{equation}
where $cov(P, IoU_{c})$ is the covariance between $P$ and $IoU_{c}$ calculated as follows:
\begin{equation}
    cov(P, IoU^{(c)}) = \sum_{i \in I} (p_{i}-\mu_{p})(IoU^{(c)}_{i}-\mu_{IoU^{(c)}}).
\end{equation}
and $\sigma_{P}$ and $\sigma_{IoU^{(c)}}$ are the standard deviations of $P$ and $IoU^{(c)}$ calculated as follows:
\begin{equation}
    \sigma_{P} = \sqrt{\sum_{i \in I} (p_{i}-\mu_{p})^{2}}.
\end{equation}
\begin{equation}
    \sigma_{IoU^{(c)}} = \sqrt{\sum_{i \in I} (IoU^{(c)}_{i}-\mu_{IoU^{(c)}})^{2}}.
\end{equation}
where mean of $P$ is denoted by $\mu_{p}$ and of $IoU^{(c)}$ by $\mu_{IoU^{(c)}}$.
The correlation coefficient is positive when the performance improves as synthetic data quality increases and is negative when the performance downgrades as synthetic data quality increases.

We employ 3D point cloud semantic segmentation networks PointNet++ \citep{PointNet} and KPConv \citep{thomas2019kpconv}, both established in 3D semantic segmentation tasks and in the geospatial community, due to their reliable performances and their direct operations on unordered point clouds: 
Both methods support training on mere 3D points. Their architectures are designed to utilize spatial relationships between neighboring points to identify local 3D structures. 
While they both leverage the spatial property of the point clouds, PointNet++ extracts the geospatial feature with a set of abstraction levels and multi-layer perceptrons stacking up in a hierarchical structure, and KPConv leverages 3D kernels to aggregate local neighborhood information. 
An advantage of using KPConv is that it provides an effective batch selection pipeline that can directly operate on a significant number of points in a point cloud, thus eliminating the need to handcraft a split of large point cloud files into separate scenes beforehand.

\section{Experiments}
\label{sec:experiments}
We compared the synthetic point cloud with its real-world counterpart to evaluate the domain gap between the two representations. 
The synthetic point cloud covers the same area as the real-world point cloud since it was simulated in the semantic 3D environment model.
In turn, the semantic model was manually created utilizing the same real-world point cloud.
The setup ensures minimum divergence between both point cloud representations in terms of temporal changes in the environment.

\begin{figure*}[htb]
    \centering
    \includegraphics[width=.9\linewidth]{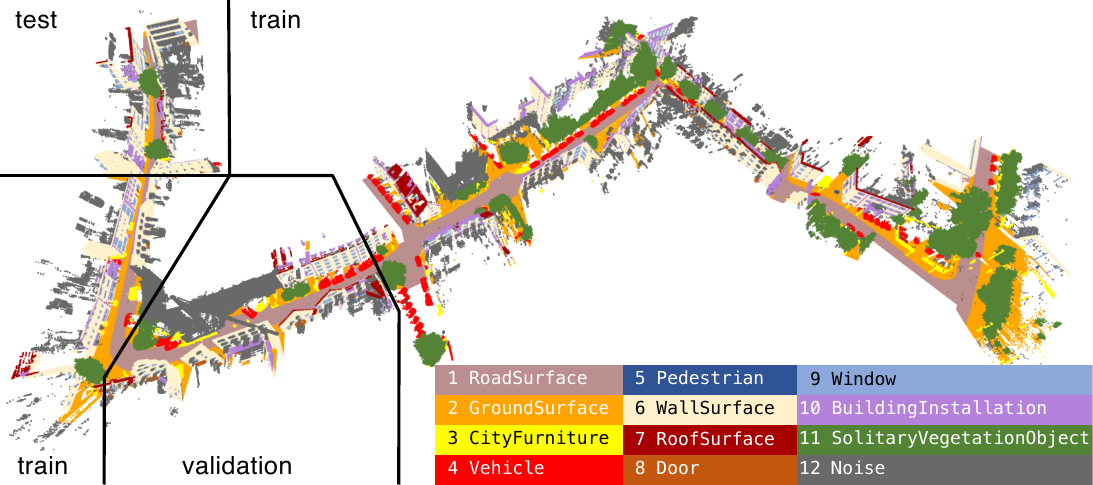}
    \caption{Data split of real-world point clouds with street lengths of approx.\ 510 meters for training, 77 meters for validation, and 85 meters for testing.}
    \label{fig:data-split}
\end{figure*}

We evaluated the synthetic point cloud generated by our proposed approach using the stochastic and deterministic methods.
We applied the evaluation methods outlined in \autoref{sec:MeasuringDomainGap} to both real-world and synthetic point clouds, analyzing the differences. 
In order to evaluate the domain gap with the stochastic approach, we conducted data splits to synthetic and real-world data, which are illustrated in \autoref{fig:data-split}.
The complete implementation of the experiments can be found in our repository, which is provided in the code availability statement. 


\subsection{Real-world data acquisition}
Our real-world point cloud dataset is captured along a designated route within the Ingolstadt city center.
The dataset covers an approximate area of 400 meters by 250 meters and includes detailed information on over 50 distinct constructions. 
The architectural styles of these buildings are commonly seen throughout Germany. 
\subsubsection{Mobile mapping point clouds}
\label{sec:MLSreal}
%
The \gls{MLS} point cloud was acquired by 3D Mapping Solutions and their in-house mobile mapping platform, which geo-referencing was supported by the German SAPOS RTK system.
The platform was mounted on a mini-van and drove while collecting the data in the city center of Ingolstadt, Germany.
Approximately 100M points with density up to 3000 $pts/m^{2}$ were recorded, whereby the setup ensured their relative accuracy in the range of 1-3 cm~\citep{mappingSolutions2023Moses}.

To ensure high-quality labeling with limited resources, we implemented a four-stage process.
First, we employed connected-component-based geometric classification to split up the original point cloud into N subsets of point clouds, each of which contains points that are geometrically close to each other while having some distance from other groups.
Secondly, we performed ground segmentation to differentiate the ground planes, such as roads, pavements, and soil from 3D objects with height.
This step yielded a significant increase in accuracy due to the prior removal of separable objects and noise points.
Next, we manually labeled the point clouds employing a polyline segmentation tool and assigned each point cloud to a numerical value corresponding to the defined classes in \autoref{subsec:semanticLabelsMapping}.
Finally, we merged all labeled subsets and resolved duplicates into one point cloud dataset.

\subsubsection{Semantic 3D city model}
\label{sec:sem3DEnvironmentModel}
\begin{figure*}[htb]
    \centering
    \includegraphics[width=.7\linewidth]{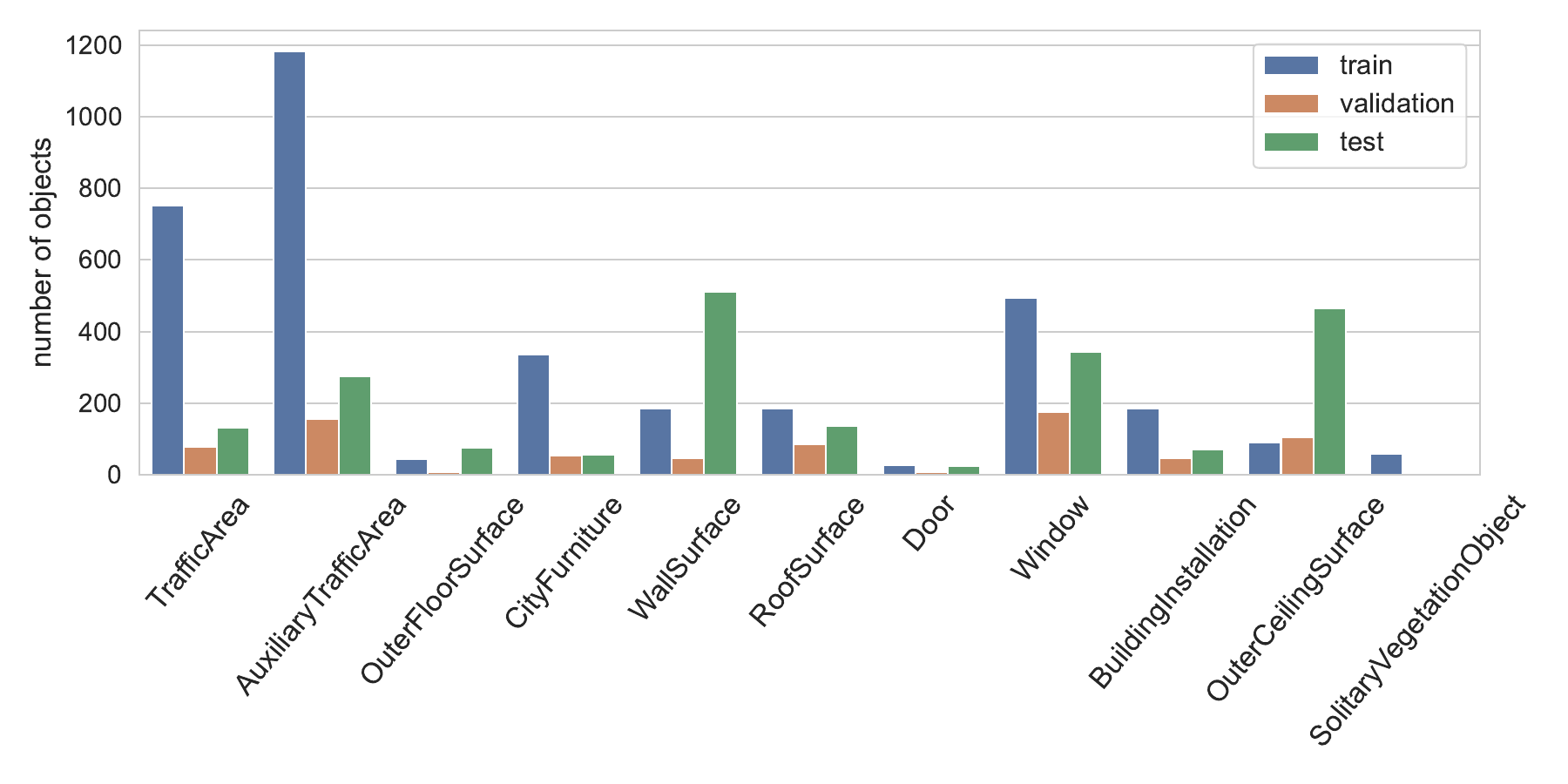}
    \caption{Number of object instances per class in the train, test, and validation CityGML dataset.}
    \label{fig:object_instances_per_class_distribution_citygml}
\end{figure*}

For our experiments, we acquired the semantic 3D city model comprising of \acrshort{LOD}3 buildings and the road environment representing the city center of Ingolstadt in Germany.\footnote{The OpenDRIVE and CityGML datasets used for the experiments are available under \url{https://github.com/savenow/lod3-road-space-models}}
\autoref{fig:object_instances_per_class_distribution_citygml} lists the number of objects per class of our semantic environment model in CityGML.
These models served as inputs for our synthetic data generation after applying our method described in \autoref{subsec:SemModelGeneration}.
Notably, these models were created manually based on the \gls{MLS} point cloud described in the previous \autoref{sec:MLSreal}, thus avoiding discrepancies caused by  different point clouds obtained in different epochs.
In our experiments, we used version 0.9.14 of the submicroscopic driving simulator CARLA as our simulation testbed, which is based on the Unreal Engine version 4 \citep{unrealengine}.

Even though we have acquired highly detailed 3D city models, some objects only approximated the real-world representation.
For example, our pipeline did not use specific methods to process complex meshes, such as trees and bushes, and thus, these elements lack geometric accuracy.
Despite the comprehensive \gls{LOD}3 models, very fine geometric structures of the real world are still generalized, such as the surface structure of doors or facade decorations.

\subsection{Synthetic Semantic Point Cloud}
\begin{figure*}[htb]
    \centering
    \includegraphics[width=0.7\linewidth]{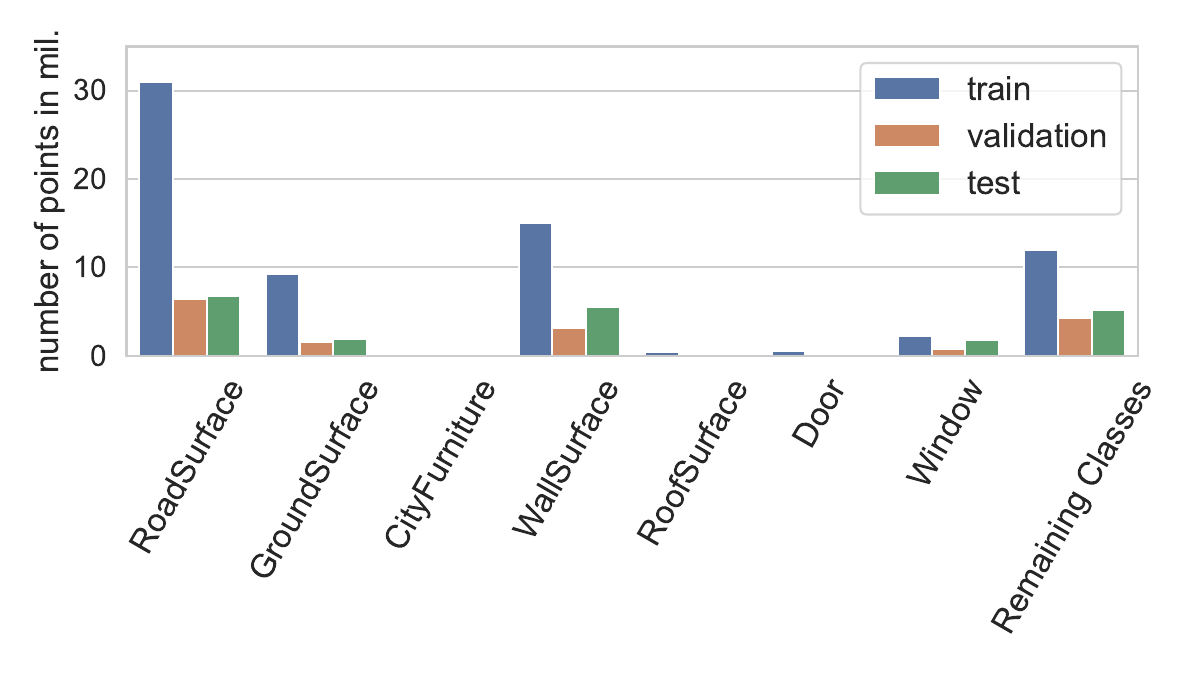}
    \includegraphics[width=0.7\linewidth]{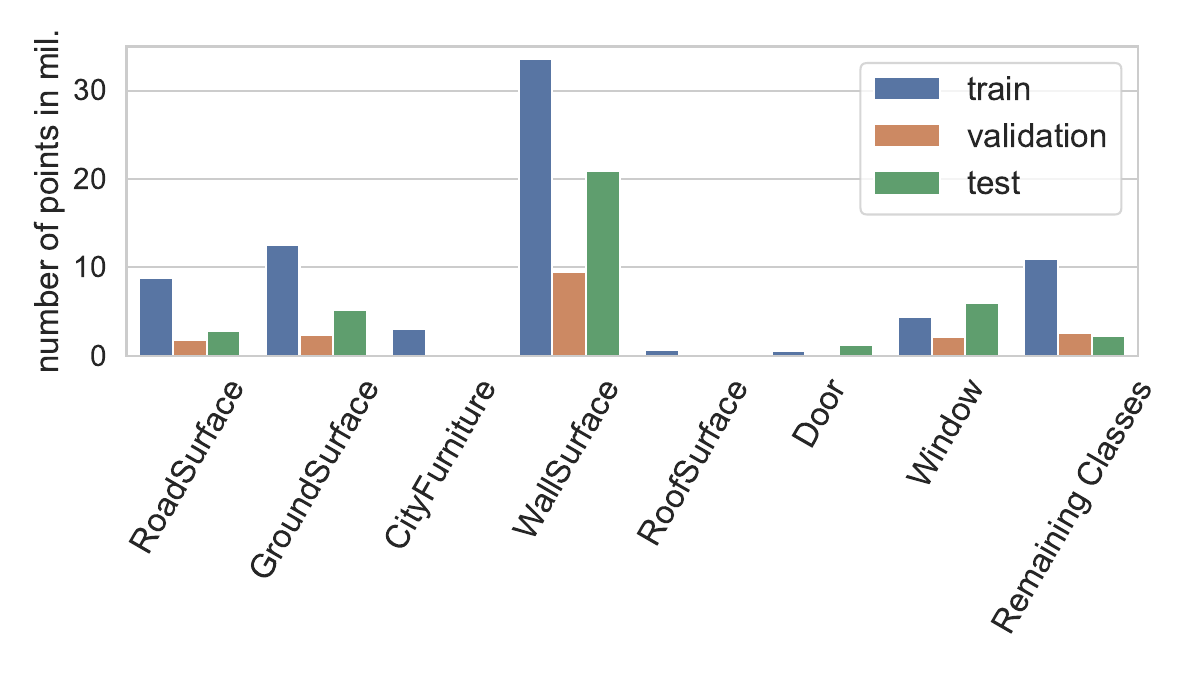}
    \caption{Number of points per object class for real-world  (left) and synthetic data (right).} 
        \label{fig:points_per_class_distribution}
\end{figure*}

To validate our method, we set up a synthetic data generation pipeline in our experiment utilizing the methodology presented in \autoref{subsec:SemModelGeneration} and created a synthetic dataset with 100M points, which yields approximately the same ratio as our real-world point cloud.
The distribution of the synthetic points compared to the real-world points per class and dataset split is depicted in \autoref{fig:points_per_class_distribution}.
%
%

\begin{figure*}[htb]
    \centering
    \includegraphics[width=0.48\linewidth]{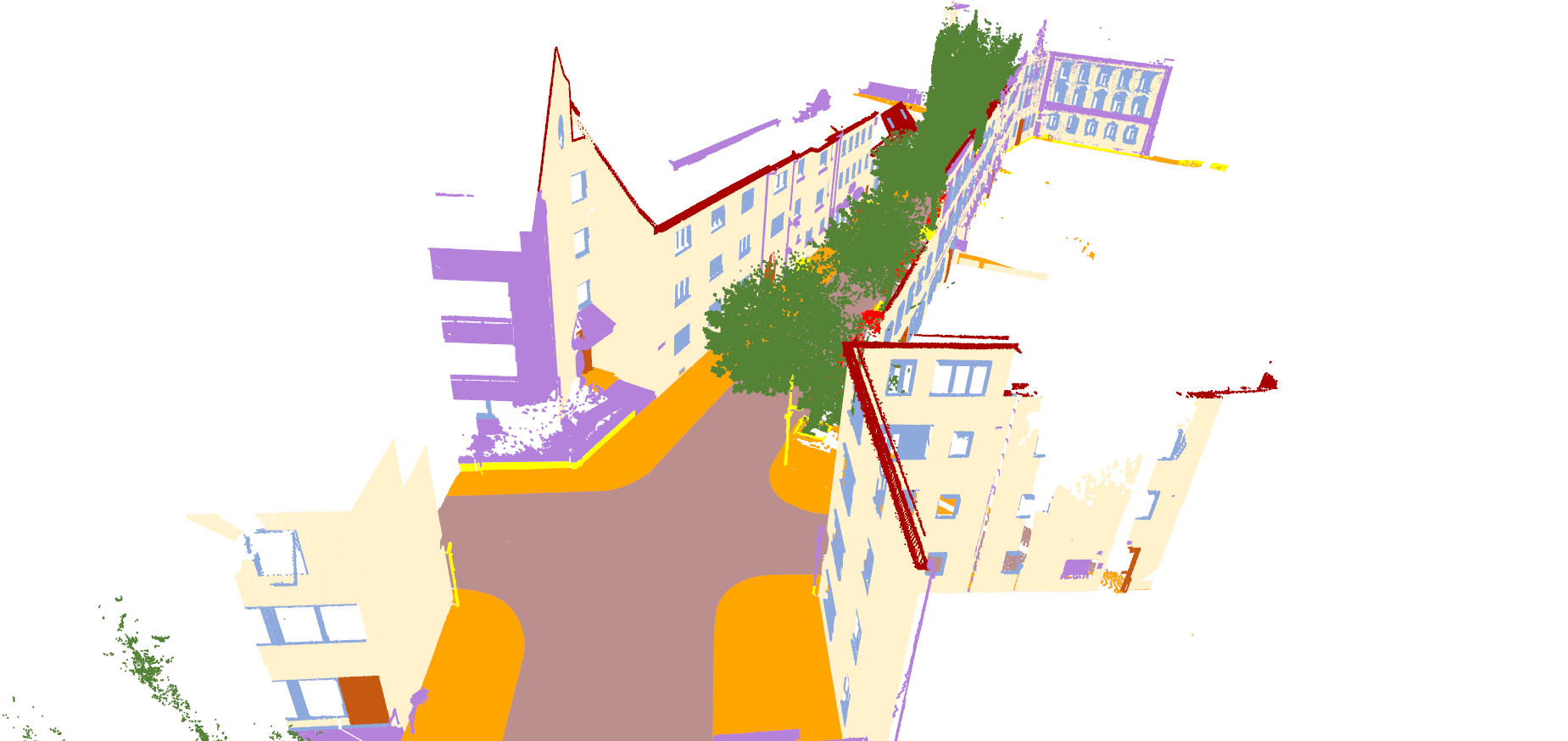}
    \includegraphics[width=0.48\linewidth]{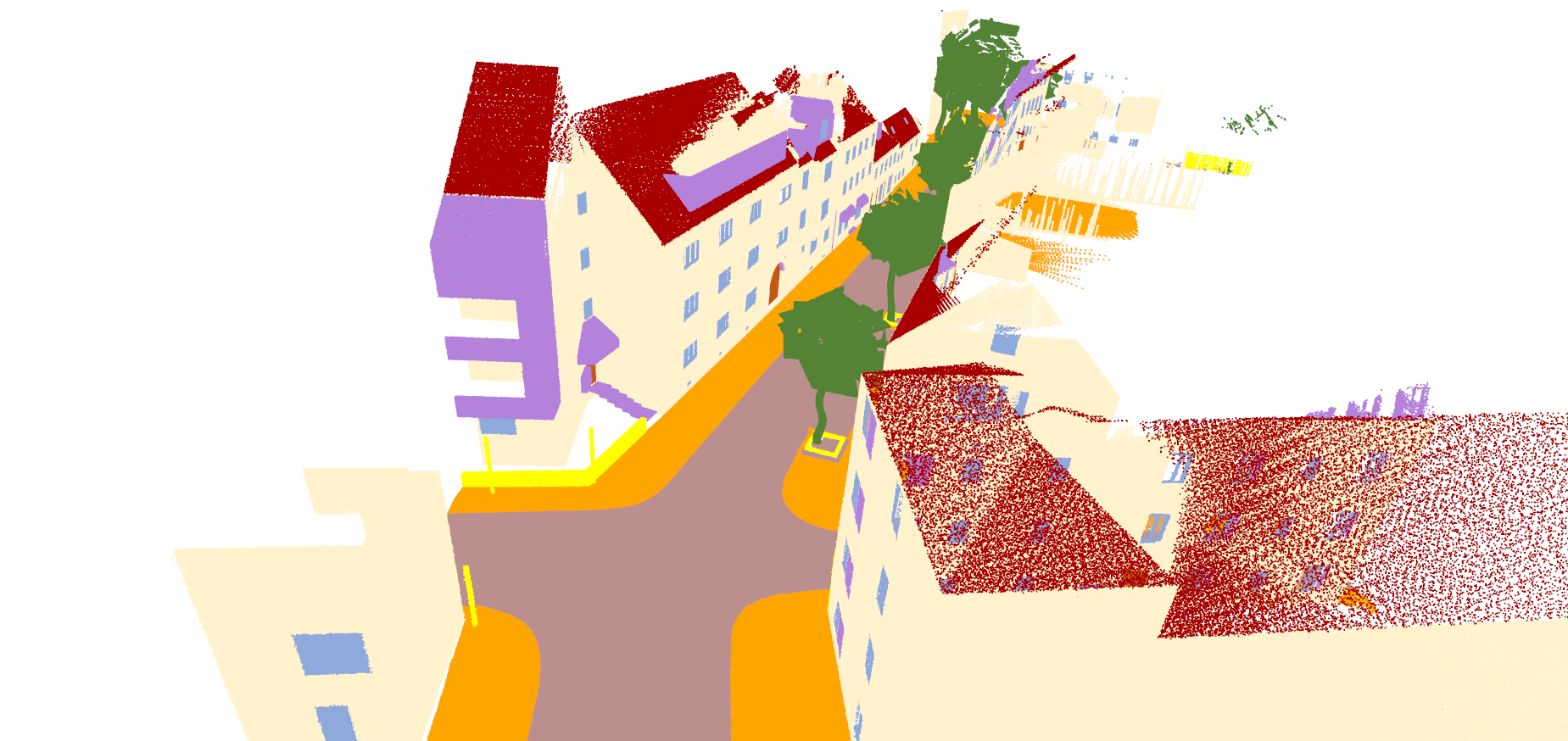}
    \caption{Comparison of the real-world (left) and synthetic point clouds (right) highlighting the discrepancies due to abstractions in the environment and sensor model.}
    \label{fig:comparison_real_synthetic_pc}
\end{figure*}

\begin{figure}[htb]
    \centering
    \includegraphics[width=\linewidth]{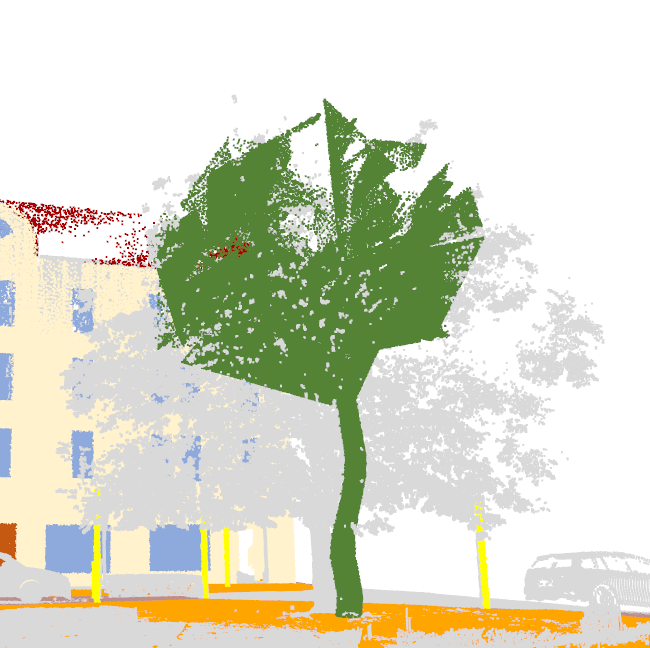}
    \caption{Synthetic point cloud of a tree in green and its real-world counterpart in gray. The synthetic point cloud representation contains all the elements of a tree, such as branches and leaves, but the geometric shapes are different.}
    \label{fig:synthetic_tree}
\end{figure}

Since the ground-truth vehicle trajectory is not disclosed by the surveying company along with the real-world point cloud, we approximated a plausible ego-vehicle trajectory. 
To avoid robotic driving behavior, we designed the control of the ego vehicle as a video game setup and drove manually.
The simulated \gls{LiDAR} mimicked real-world sensors, capturing scans at realistic rates with post-processing as mentioned in \autoref{subsec:simulatingLaser}. 
The physical interactions between artifacts, such as collision and gravitational forces, were deactivated due to unexpected behavior from vehicle-obstacle interactions. 
%
%
\autoref{fig:comparison_real_synthetic_pc} illustrates the real-world point cloud of a street section side-by-side with the synthetically generated counterpart.
Around 100M points were generated for the entire area, which is comparable to the acquired real point cloud.
As we mentioned in \autoref{sec:sem3DEnvironmentModel}, some 3D objects approximated the real-world point cloud representation. This phenomenon is shown in \autoref{fig:synthetic_tree}.

\subsection{Evaluation of the Deterministic Approach}
\label{subsec:evaluationDeterministic}
\subsubsection{Parameters}
Since our real-world point clouds are georeferenced, we first translated them into the same local reference frame as the synthetic point cloud.
We calculated the \acrshort{mIoU} utilizing a voxel structure and chose a voxel size of $l_{voxel} = 0.5\text{ m}$, following the usual divergence in the geo-registration of mobile mapping point clouds \citep{zhu_tum-mls-2016_2020,wysocki2023scan2lod3}. 
The voxel size was chosen empirically considering point cloud density and experiments performed under arbitrary translation values listed in \autoref{tab:voxel_shifts}. 
\begin{table}[htbp]
    \centering
    \caption{Metric sensitivity with the selected offset distances under constant voxel size in the local reference frame.}
    \label{tab:voxel_shifts}
    \footnotesize
    \begin{tabular}{@{} l c c c @{}}
        \toprule
         & \multicolumn{3}{c}{\textbf{Offset}} \\
        \cmidrule{2-4}
         & 0.0\,m & 0.1\,m & 0.3\,m \\
        \midrule
        $m_{DoGSS-PCL}{(\mathcal{R}, \mathcal{S})}$ & 0.09 & 0.51 & 0.59 \\
        $d(\mathcal{R}, \mathcal{S})$ & 0.15 & 0.46 & 0.55\\
        $d_{MM3C2}$ & 0.04 & 0.22 & 0.35\\
        $d_{C2C}$ & 0.31 & 0.80 & 0.85\\
        $mIoU$ & 29.07\% & 3.10\% & 2.50\%\\
        \bottomrule
    \end{tabular}
\end{table}

The class weights for \acrshort{mIoU} and $d_{MM3C2}$ were chosen to emphasize building-related classes and are shown in \autoref{tab:class_weights}. 
We considered walls essential as they constitute the primary geometry of buildings and, thus, the city's geometry.
We decided to perform the comparison on well-represented static objects, as listed in \autoref{tab:class_weights}.
In consequence, the comparison omits the dynamic objects not present in the virtual model, such as \texttt{Vehicle}, \texttt{Pedestrian}; also, largely simplified objects, such as \texttt{RoadSurface}, \texttt{SolitaryVegetationObject}; as well as noise represented by the \texttt{Noise} class.
\begin{table}[htbp]
    \centering
    \caption{Weights per class used for weighted average computation \acrshort{mIoU} and $d_{MM3C2}$, chosen based on the heuristic and class importance.}
    \label{tab:class_weights}
    \footnotesize
    \begin{tabular}{@{} l c  @{}}
        \toprule
        \textbf{Class} & \textbf{Weight} \\
        \midrule
        \texttt{CityFurniture} & 0.1 \\
        \texttt{GroundSurface} & 0.1 \\
        \texttt{WallSurface} & 0.2 \\
        \texttt{RoofSurface} & 0.15 \\
        \texttt{Door} & 0.15 \\
        \texttt{Window} & 0.15 \\
        \texttt{BuildingInstallation} & 0.15 \\
        \bottomrule
    \end{tabular}
\end{table}
%
%
For the metric $m_{\mathcal{R}}(\mathcal{S})$, we chose the weights and consider that by construction, even acceptable \acrshort{mIoU} values will lead to high scores. 
To have a balance between $d_{MM3C2}$ and $f_{mIoU}$, we thus empirically selected $\lambda_1 = 0.6$, $\lambda_2 = 0.3$, and $\lambda_3 = 0.1$ for $f_{mIoU}$. 
For the bounded growth rate, we empirically chose $\alpha = -0.2$ as it led to a wide range of values in the range $[0, 1]$ being used with the data for this paper. 

As the metric proposed indicates only the relative fit of the synthetic point clouds, the parameters for the metric can be freely tuned by the user as long as they remain constant for all synthetic point clouds that are compared at a time.
\subsubsection{Results}
%
To better demonstrate the sensitivity of our deterministic approach, we randomly translated the synthetic point cloud by around $1$ m and evaluated our \gls{DoGSS-PCL} metric on both the original and translated versions. The results are presented in \autoref{tab:distance_metric_results}.
\begin{table}[htbp]
    \centering
    \caption{Distance and metric results.}
    \label{tab:distance_metric_results}
    \begin{tabular}{@{} l c c @{}}
        \toprule
        & Synthetic & Synthetic (translated) \\
        \midrule
        \textbf{$m_{DoGSS-PCL}{(\mathcal{R}, \mathcal{S})}$}  & $0.09$ & $0.73$\\
        \textbf{$d(\mathcal{R}, \mathcal{S})$} & $0.15$ & $1.14$ \\
        \midrule
        $d_{MM3C2}$  & $0.04$ & $1.01$\\
        $d_{C2C}$  & $0.31$ & $1.34$\\
        $mIoU$ & $0.29$ & $0.02$ \\
        $f_{mIoU}$  & $3.44$ & $54.79$\\
        \bottomrule
    \end{tabular}
\end{table}

With the synthetic point cloud $\mathcal{S}$, the distances were in a cm range, while the \acrshort{mIoU} was in a low two-digit span. 
This led to a relatively small $IoU^{(c)}$ factor of $3.44$. 
Accordingly, the original synthetic point cloud performed well on our custom metric with a value of $m_{DoGSS-PCL}{(\mathcal{R}, \mathcal{S})} = 0.09$, which was much closer to $0$ than to $1$ and thus constituted a good value given the metric parameters we chose.

For the translated synthetic point cloud, the \acrshort{mIoU} score decreased by one order of magnitude. 
This decrease was influenced by the voxel size $l_{voxel} = 0.5\text{ m} < 1.14 = d({\mathcal{R}, \mathcal{S})}$ being smaller than the cloud distance.
Essentially, the resolution of the $IoU^{(c)}$ calculation was higher than the distance of points, which led to low $IoU^{(c)}$ scores. 
In contrast, with the non-translated synthetic point cloud $d ({\mathcal{R}, \mathcal{S})} = 0.15 < 0.5 = l_{voxel}$, thus the $IoU^{(c)}$ was significantly higher. 
The low \acrshort{mIoU} as well as the higher distances led to a metric score of $m_{DoGSS-PCL}{(\mathcal{R}, \mathcal{S})} = 0.73$, which was closer to $1$ than to $0$ and much higher than the translated score of $m_{DoGSS-PCL}{(\mathcal{R}, \mathcal{S})} = 0.09$. 

\subsection{Evaluation of the Stochastic Approach}
\label{subsec:evaluationStochasticApproach}
\subsubsection{Parameters}
\begin{figure}
  \centering
  \begin{tabular}{@{}c@{}}
    \includegraphics[width=.75\linewidth]{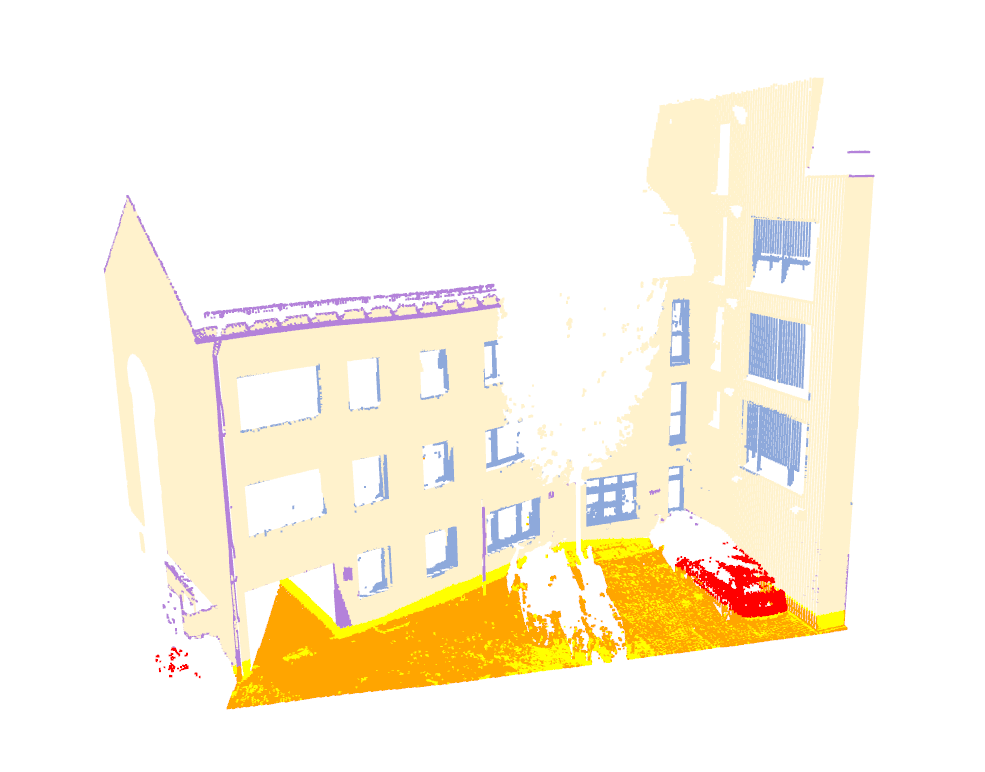} 
    \label{fig:real75}\\
    \small (a) real-world points
  \end{tabular}
  \vspace{\floatsep}
  \begin{tabular}{@{}c@{}}
    \includegraphics[width=.75\linewidth]{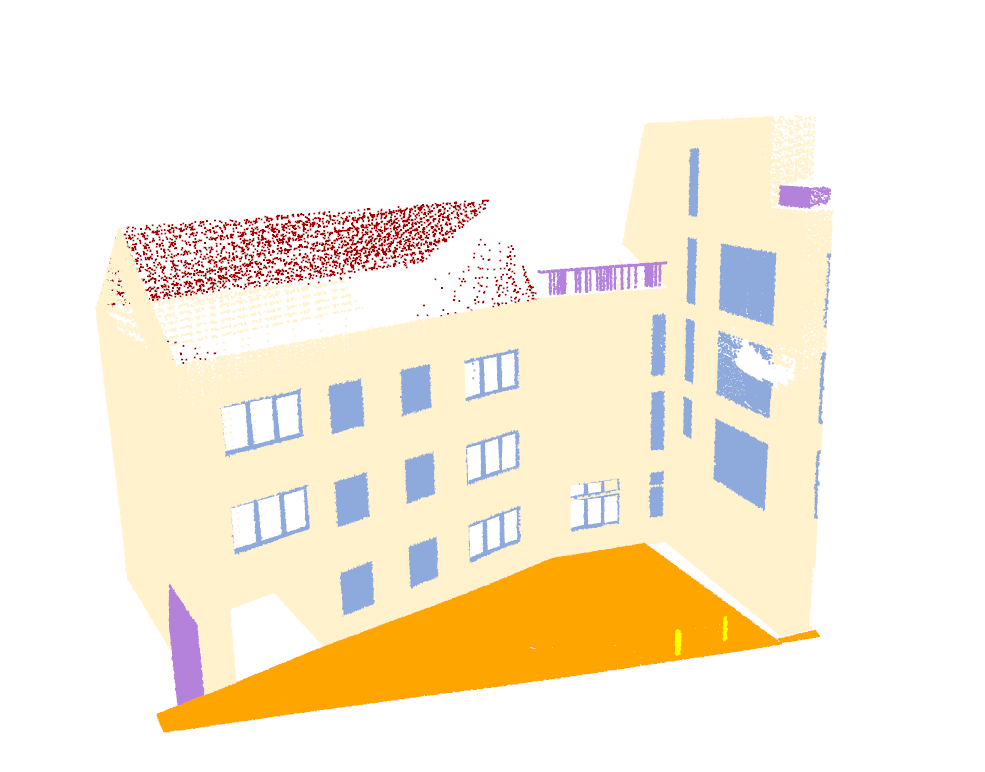}
    \label{fig:synth25}\\
    \small (b) synthetic points
  \end{tabular}
  \vspace{\floatsep}
  \begin{tabular}{@{}c@{}}
    \includegraphics[width=.75\linewidth]{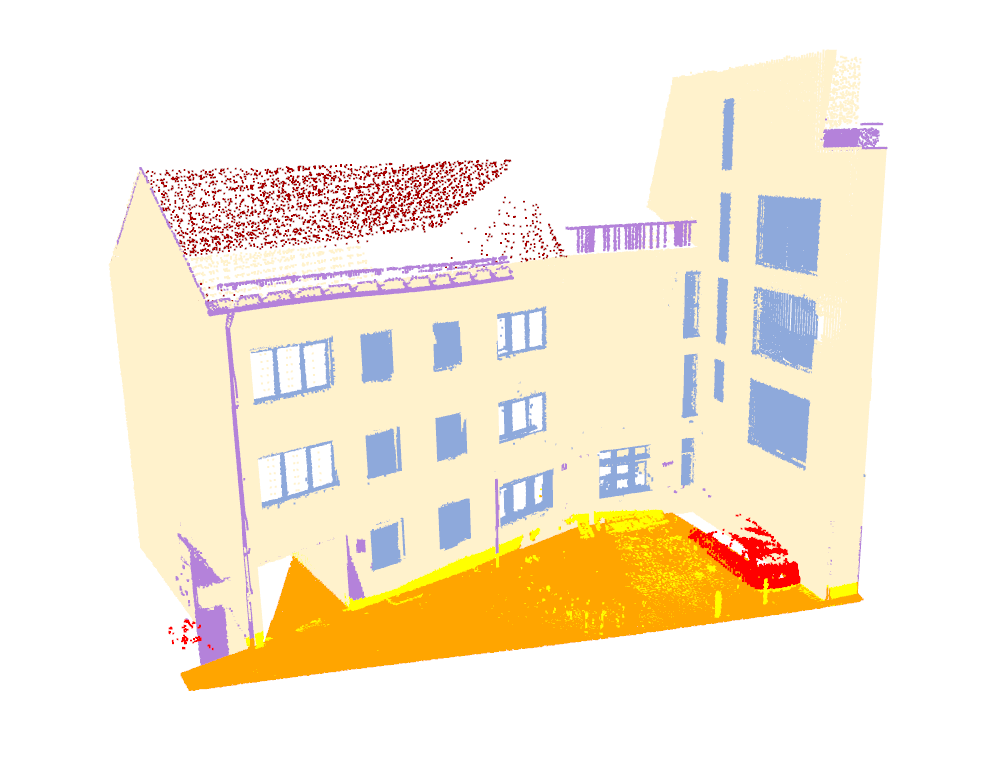}
    \label{fig:mix}\\
    \small (c) 50\%-50\% mixture
  \end{tabular}
  \caption{Real-world and synthetic point cloud mixture.}
  \label{fig:data-mixture}
\end{figure}

We designed the following training sets of the real-to-synthetic ratio: 100\%-0\%, 75\%-25\%, 50\%-50\%, 25\%-75\%, and 0\%-100\%. 
The former percentage indicates the quantity ratio of the real-world point cloud to the set, and the latter is the quantity ratio of the synthetic point cloud to the set. 
For each set, we randomly sourced points from the real-world domain and synthetic domain, respectively, to the desired proportions, then concatenated both point clouds into a single point cloud. 
We used the training split area specified in~\autoref{fig:data-split} to establish the training sets. 
We evaluated the results on the test split of the real-world point clouds. ~\autoref{fig:data-mixture} visualizes the result for a 50\%-50\% mixture of real-world (\autoref{fig:data-mixture}a, blue color palette) and synthetic data (\autoref{fig:data-mixture}b, red color palette).
~\autoref{fig:data-mixture}c shows the case where the synthetic point cloud complemented a region of the wall surface that was occluded by the tree.

To train PointNet++ with our datasets, we split our point clouds into sub-scenes. 
In each iteration, an input scene served as a batch and passed through the model. 
Since KPConv randomly samples input batches directly from large point clouds, it eliminated the need to split the point clouds into small scenes beforehand.
In each iteration, several points and their spherical neighborhoods were queried, augmented, and stacked together as an input batch. 
To sample different regions of the clouds more evenly, selected points in an iteration were made to be chosen less likely in the subsequent iterations.

We trained both networks with point coordinates without additional features such as color or normal vectors. 
The model structures, optimizers, losses, and hyperparameters, such as the neighborhood querying radius of PointNet++ and KPConv, were inspired by and  followed the ones used in the original works. 
The models were evaluated on the unseen test real-world point cloud data, as shown in \autoref{tab:kpconv-class-wise-iou-test}, \autoref{tab:pnpp-class-wise-iou-test}, and in \autoref{fig:ratios}.

\subsubsection{Results}

We observe that KPConv outperforms PointNet++, corroborating the current research consensus, e.g., as reported in the Paris-Carla-3D dataset \citep{carla_paris}.

The performance difference between the two models is substantial, as found in~\autoref{tab:kpconv-class-wise-iou-test} and~\autoref{tab:pnpp-class-wise-iou-test}.
Hence, when analyzing the domain gap between synthetic and real-world data, we focused mainly on the results of KPConv as it provided us with a much more comprehensive evaluation than the results of PointNet++.

In general, the $mIoU$ decreases as the ratio of synthetic data increases with $corr. = -0.8$ for KPConv and $corr. = -0.62$ for PointNet++.
Nevertheless, with 50\% of synthetic data, KPConv still performs close to the results of training KPConv with 100\% real-world data. 
Generally, the performance worsens as the ratio of synthetic point clouds increases. An exception is the \texttt{RoofSurface} class, which holds a positive correlation coefficient of $0.60$ with KPConv and $0.12$ with PointNet++. 
Unlike the other classes, the $IoU^{(c)}$ of \texttt{RoofSurface} object class increases as the synthetic point clouds increase.

\begin{figure*}[htb]
    \centering
    \includegraphics[width=\linewidth]{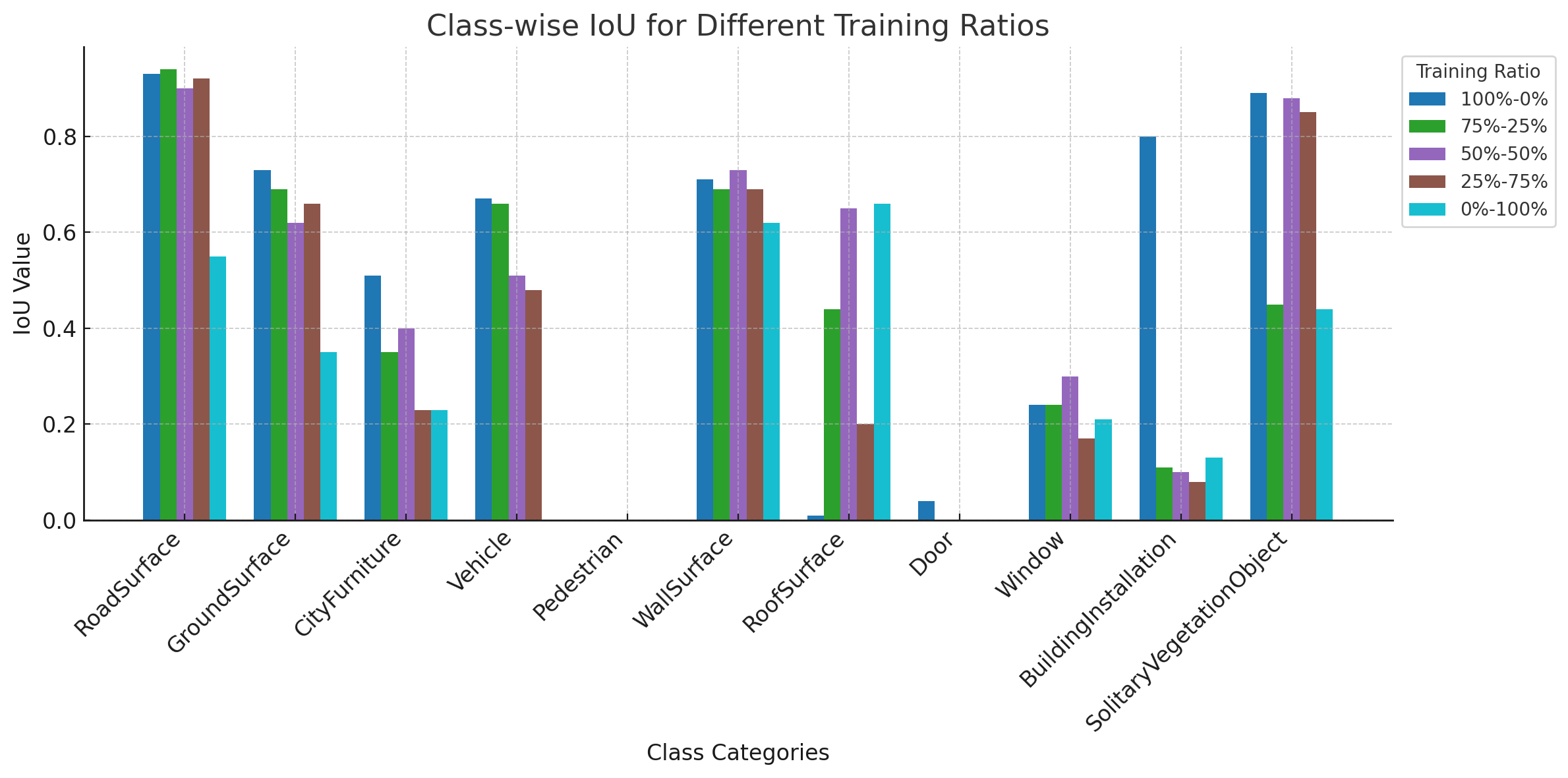}
    \caption{KPConv results (\autoref{tab:kpconv-class-wise-iou-test}) visualized on a bar chart (real vs synthetic). The analysis corroborates the non-linear correlation between the mixed ratio setups with a large dependency on the object in question. The intuitive hypothesis that the performance shall have the highest scores holds for the most classes, with the striking example of \texttt{BuildingInstallation} (drop by 70\%). Remarkably, for such classes as \texttt{Window}, the supplementary synthetic data can increase the generalization capabilities, even outperforming real-only input, presumably due to laser glass-penetration and noise effects rendering real-world point clouds very sparse, in contrast to 3D simulated windows yielding complete windows geometries.}
    \label{fig:ratios}
\end{figure*}

We provide the class-wise $IoU^{(c)}$, \gls{mIoU}, and the correlation coefficient results in ~\autoref{tab:kpconv-class-wise-iou-test}, ~\autoref{tab:pnpp-class-wise-iou-test}, and in \autoref{fig:ratios}.

\begin{table*}[htb]
    \centering
    \caption{KPConv class-wise $IoU^{(c)}$ and \gls{mIoU} evaluated on real-world point cloud with training ratio: \textbf{real\%-synthetics\%}. The column \textbf{avg.} states the average $IoU^{(c)}$ for each class. The column \textbf{corr} displays the correlation coefficient between the class-wise $IoU^{(c)}$ trends and synthetic data ratio.}
    \label{tab:kpconv-class-wise-iou-test}
    \begin{tabular}{@{} l|c c c c c c c c @{}}
        \toprule
        & \textbf{100\%-0\%} & \textbf{75\%-25\%} & \textbf{50\%-50\%} & \textbf{25\%-75\%} & \textbf{0\%-100\%} & \textbf{avg.} & \textbf{corr.} \\
        \midrule
        \texttt{RoadSurface}          & $0.93$ & $0.94$ & $0.90$ & $0.92$ & $0.55$ & $0.85$ & $-0.70$ \\
        \texttt{GroundSurface}        & $0.73$ & $0.69$ & $0.62$ & $0.66$ & $0.35$ & $0.61$ & $-0.80$ \\
        \texttt{CityFurniture}     & $0.51$ & $0.35$ & $0.40$ & $0.23$ & $0.23$ & $0.34$ & $-0.90$ \\
        \texttt{Vehicle}              & $0.67$ & $0.66$ & $0.51$ & $0.48$ & $0.00$ & $0.46$ & $-0.90$ \\
        \texttt{Pedestrian}           & $0.00$ & $0.00$ & $0.00$ & $0.00$ & $0.00$ & $0.00$ & $-$ \\
        \texttt{WallSurface}          & $0.71$ & $0.69$ & $0.73$ & $0.69$ & $0.62$ & $0.70$ & $-0.70$ \\
        \texttt{RoofSurface}          & $0.01$ & $0.44$ & $0.65$ & $0.20$ & $0.66$ & $0.39$ & \phantom{+}$0.60$ \\
        \texttt{Door}                 & $0.04$ & $0.00$ & $0.00$ & $0.00$ & $0.00$ & $0.01$ & $-0.70$ \\
        \texttt{Window}               & $0.24$ & $0.24$ & $0.30$ & $0.17$ & $0.21$ & $0.23$ & $-0.40$ \\
        \texttt{BuildingInstallation} & $0.80$ & $0.11$ & $0.10$ & $0.08$ & $0.13$ & $0.24$ & $-0.70$ \\
        \texttt{SolitaryVegetationObject}                 & $0.89$ & $0.45$ & $0.88$ & $0.85$ & $0.44$ & $0.70$ & $-0.30$ \\
        \noalign{\smallskip}\hline\noalign{\smallskip}
        $\mathbf{mIoU}$ & $0.45$ & $0.45$ & $0.46$ & $0.39$ & $0.29$ & $0.41$ & $-0.80$ \\
    \bottomrule
    \end{tabular}
\end{table*}

\begin{table*}[htb]
    \centering
    \caption{PointNet++ class-wise $IoU^{(c)}$ and \gls{mIoU} evaluated on real-world point cloud with training ratio: real\%-synthetics\%}
    \label{tab:pnpp-class-wise-iou-test}
    \begin{tabular}{@{} l|c c c c c c c c @{}}
        \toprule
        & \textbf{100\%-0\%} & \textbf{75\%-25\%} & \textbf{50\%-50\%} & \textbf{25\%-75\%} & \textbf{0\%-100\%} & \textbf{avg.} & \textbf{corr.} \\
        \midrule
        \texttt{RoadSurface}          & $0.64$ & $0.34$ & $0.37$ & $0.42$ & $0.26$ & $0.41$ & $-0.76$ \\
        \texttt{GroundSurface}        & $0.27$ & $0.13$ & $0.16$ & $0.20$ & $0.16$ & $0.18$ & $-0.48$ \\
        \texttt{CityFurniture}     & $0.03$ & $0.00$ & $0.00$ & $0.00$ & $0.01$ & $0.01$ & $-0.50$ \\
        \texttt{Vehicle}              & $0.08$ & $0.01$ & $0.00$ & $0.00$ & $0.00$ & $0.02$ & $-0.78$ \\
        \texttt{Pedestrian}           & $0.00$ & $0.00$ & $0.00$ & $0.00$ & $0.00$ & $0.00$ & $-$ \\
        \texttt{WallSurface}          & $0.34$ & $0.20$ & $0.32$ & $0.25$ & $0.35$ & $0.29$ & $-0.20$ \\
        \texttt{RoofSurface}          & $0.00$ & $0.05$ & $0.00$ & $0.01$ & $0.03$ & $0.02$ & \phantom{+}$0.12$ \\
        \texttt{Door}                 & $0.02$ & $0.00$ & $0.02$ & $0.03$ & $0.00$ & $0.02$ & $-0.25$ \\
        \texttt{Window}               & $0.09$ & $0.05$ & $0.06$ & $0.07$ & $0.10$ & $0.07$ & $-0.23$ \\
        \texttt{BuildingInstallation} & $0.05$ & $0.00$ & $0.02$ & $0.04$ & $0.01$ & $0.03$ & $-0.39$ \\
        \texttt{SolitaryVegetationObject}                 & $0.16$ & $0.14$ & $0.08$ & $0.08$ & $0.07$ & $0.11$ & $-0.90$ \\
        \noalign{\smallskip}\hline\noalign{\smallskip}
        $\mathbf{mIoU}$ & $0.15$ & $0.08$ & $0.10$ & $0.10$ & $0.09$ & $0.10$ & $-0.62$ \\
    \bottomrule
    \end{tabular}
\end{table*}

\section{Discussion}

\subsection{Deterministic Approach}
The \emph{DoGSS-PCL} metric we propose evaluates the quality of synthetic point clouds by combining physical and semantic distances into a single score \emph{DoGSS-PCL} $\in [0,1]$. 
By capturing both global and semantic grouped distance, the metrics can capture both global changes and semantic mismatches. We showcase its effectiveness by comparing well-fitting semantic point clouds with translated ones (\autoref{tab:distance_metric_results}). 
As expected, the well-fitting version achieves a significantly lower \emph{DoGSS-PCL} score, highlighting its capturing capabilities of both global structure and semantic fidelity. 
Notably, \emph{DoGSS-PCL} provides a versatile evaluation, measuring global and semantic-based geometric mismatch, while also incorporating $f_mIoU$ to assess semantic accuracy, particularly for underrepresented classes.

The synthetic point cloud is well aligned geometrically and harmonized semantically, which could be seen from our metric \emph{DoGSS-PCL}. 
In detail, the recorded global mismatch is quite small with mean bi-directional $d_{C2C} \approx 0.1$. The distance becomes even smaller when we consider semantic information, where most classes maintain a small stable deviation between real-world and synthetic point clouds with a mean of $0.04$ and a standard deviation around $0.5$.
$d_{MMC2C}$ can also capture the difference in geometry of the \texttt{RoofSurface}, \texttt{GroundSurface}, and \texttt{BuildingInstallation}. 
However, this would only be reflected by an increase of the mean distance to $0.08$  and standard derivation to $0.14$. Unfortunately, these increases would make a minor contribution to our final result due to the imbalance in the number of points. 
\begin{table}[htb]
    \centering
    \caption{Class-wise $IoU^{(c)}$, where classes \texttt{Pedestrian}, \texttt{Vehicle} are not taken into account since dynamic objects were not present in the synthetic dataset.}
    \begin{tabular}{@{} ll @{}}
    \toprule
    Class & \% \\
    \midrule
    \texttt{RoadSurface} (1): & 66.01 \\
    \texttt{GroundSurface} (2): & 34.75 \\
    \texttt{WallSurface} (6): & 29.31 \\
    \texttt{RoofSurface} (7): & 12.67 \\
    \texttt{Doors} (8): & 42.08 \\
    \texttt{Windows} (9): & 22.53 \\
    \texttt{BuildingInstallations} (10): & 10.29 \\
    \bottomrule
    \end{tabular}
    \label{tab:iou_per_class}
\end{table}

While metrics such as $d_{C2C}$ and $d_{MMC2C}$ effectively capture the overall discrepancy between two point clouds, they fail to account for subtle variations in geometric characteristics. These geometric details are critical when comparing synthetic and real-world point clouds, as even minor deviations in surface smoothness from classes such as road, roof, and ground would make all the difference here. To address this limitation, \emph{DoGSS-PCL} computes additionally $f_{mIoU}$, which specifically targets geometric differences between point clouds of objects. 

Our metric reveals a limitation in the synthetic point clouds: There exists a significant deviation in geometry for classes such as \texttt{CityFurniture} and \texttt{BuldingInstallation}. This is reflected prominently in our metric. \emph{DoGSS-PCL} permits a versatile evaluation of the point clouds, avoiding the complexities of comparing multiple independent measures simultaneously. 
In the synthetic data, classes \texttt{Window} and \texttt{Road} have been simulated quite accurately, corresponding to $40\%$ and $60\%$ $IoU^{(c)}$. 
Conversely, classes with more elevated geometric variability, such as \texttt{GroundSurface} and \texttt{Door}, can only achieve approximately $30\%$. 
Highly detailed classes such as \texttt{BuldingInstallation} scored low $IoU^{(c)}$ with around $10\%$ because \texttt{BuldingInstallation}'s geometry was simplified significantly in our model into simple shapes, which has been well-detected by \emph{DoGSS-PCL}. 
The details $IoU^{(c)}$ for each class can be found in \autoref{tab:iou_per_class}. 

Overall, synthetic data exhibits good global and semantic alignment with its real-world counterpart. While the synthetic and real-world point clouds are well-aligned globally, there is still room for improvement in modeling fine-grained geometric details, especially with classes \texttt{GroundSurface} and \texttt{BuldingInstallation}. These geometric discrepancies can significantly impact the performance of downstream models, as we will explore in the following section.

\subsection{Stochastic Approach}

We discuss the pros and cons of training the model with our synthetic data based on the observations of the segmentation performances on the real-world point cloud test set. As stated previously, we focused mainly on the results of KPConv.

One of the flaws of synthetic data is their lack of distinct features on planar surfaces, which often leads to under-segmentation.
This issue was corroborated by our experiments, as, for example, objects that are geometrically represented by surfaces, such as \texttt{RoadSurface} and \texttt{GroundSurface} objects, performed gradually worse with the increase of synthetic data, with correlation coefficients of $-0.7$ and $-0.8$, respectively. 

Moreover, from~\autoref{tab:kpconv-misclassified}, we observe that \texttt{RoadSurface} and \texttt{GroundSurface} objects are most easily misclassified and confused with each other owing to limited distinct features.
In the real world, instead of flat surfaces without any spectral information, \texttt{RoadSurface} and \texttt{GroundSurface} objects contain additional low-level (e.g., distinct edge color) and high-level (e.g., spatial relation to other urban elements) features, allowing humans and classifiers to distinguish between them. 
In synthetic data, these significant features disappear, simplifying both classes into flat, neighboring surfaces, rendering them prone to misclassification.
Potentially, such effects can be mitigated by adjusting neural networks' receptive fields.
\begin{table*}[htb]
    \centering
    \caption{Most misclassified classes. For example, GroundSurface appears most frequently among the false positives of RoadSurface, with $95\%$ of appearance.}
    \label{tab:kpconv-misclassified}
    \begin{tabular}{@{} llc @{}}
        \toprule
        \textbf{Wrong Prediction} & \textbf{Most Frequent Class} & \textbf{Proportion} \\
        \midrule
        \texttt{RoadSurface} & \texttt{GroundSurface} & $95\%$ \\
        \texttt{GroundSurface} & \texttt{RoadSurface} & $48\%$ \\
        \texttt{CityFurniture} & \texttt{GroundSurface} & $32\%$ \\
        \texttt{Vehicle} & \texttt{CityFurniture} & $75\%$ \\
        \texttt{Pedestrian} & \texttt{-} & $-$ \\
        \texttt{WallSurface} & \texttt{Window} & $44\%$ \\
        \texttt{RoofSurface} & \texttt{BuildingInstallation} & $87\%$ \\
        \texttt{Door} & \texttt{WallSurface} & $52\%$\\
        \texttt{Window} & \texttt{BuildingInstallation} & $71\%$ \\
        \texttt{BuildingInstallation} & \texttt{WallSurface} & $48\%$ \\
        \texttt{SolitaryVegetationObject} & \texttt{BuildingInstallation} & $42\%$ \\
        \bottomrule
    \end{tabular}
\end{table*}

Another drawback of the virtual models we observed was the wide variety of object structures in the \texttt{BuildingInstallation} class, leading to model confusion. 
The objects in the \texttt{BuildingInstallation} class often consist of different parts that have a structure similar to those of other classes.
In synthetic data, \texttt{BuildingInstallation} objects often consist  of irregular shapes as well as vertical flat planes that resemble \texttt{WallSurface} objects, which is the most misclassified class (see ~\autoref{tab:kpconv-misclassified}). 
The model then fails to extract significant and consistent features for the class.

\begin{figure}[htb]
    \centering
    \includegraphics[width=1.0\linewidth]{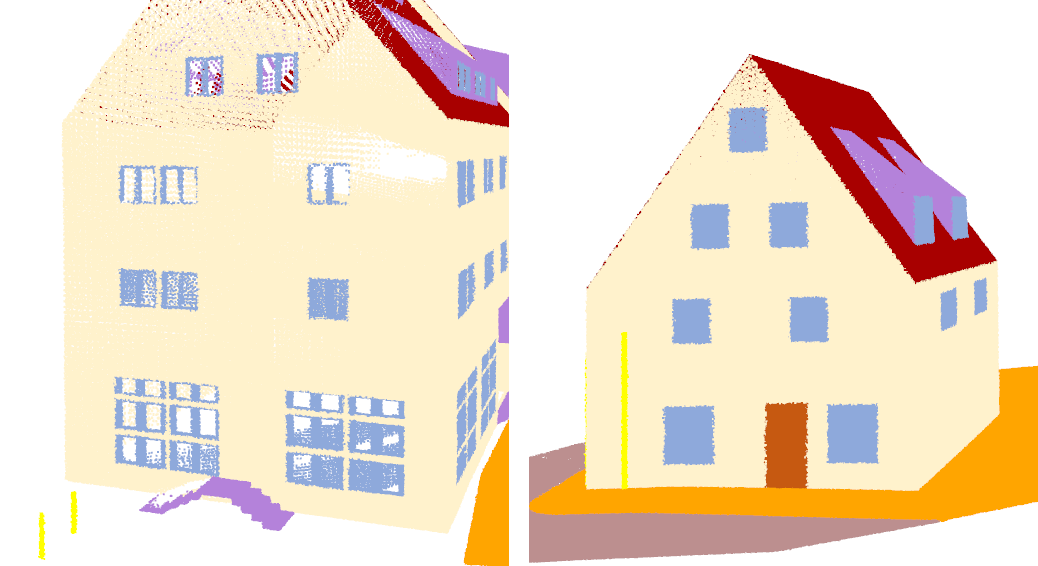}
    \caption{\texttt{Window} objects simplified into surfaces (right) and \texttt{Window} objects with complex structure (left).}
    \label{fig:window}
\end{figure}

Although surfaces in synthetic data possess problems of lack-of-feature and objects in the \texttt{BuildingInstallation} class lack significant structure, some other classes that have unique and consistent structures are, on the other hand, helpful in training. 
For example, the segmentation of class \texttt{SolitaryVegetationObject} performs well with low dependency on the ratio of synthetic data ($corr. = -0.3$), even though the tree is not accurately modeled in synthetic data. 
This phenomenon can be explained by the unique and consistent structure of \texttt{SolitaryVegetationObject} objects that other classes do not possess. 
Both real-world and synthetic data consistently represent trees as significant, spherical-like structures with a vertical trunk. 
As this structure barely appears in other classes, the model can recognize it easily without confusion. 
The identical structure of \texttt{SolitaryVegetationObject} class objects is maintained when the ratio of synthetic data accumulates, thereby providing a consistent learning source for the model.
Consequently, this consistency translates into a relatively steady performance, a pattern also observable in the \texttt{Window} class. 
Both in real-world and synthetic data, \texttt{Window} objects possess unique structures with window frames and glass surfaces fused, as shown in~\autoref{fig:window}. 
However, in synthetic data, some \texttt{Window} objects are simplified into surfaces without window frames (\autoref{fig:window}), which causes confusion between \texttt{WallSurface} objects and \texttt{Window} objects (\autoref{tab:kpconv-misclassified}).

\begin{figure}[htb]
    \centering
    \includegraphics[width=1.0\linewidth]{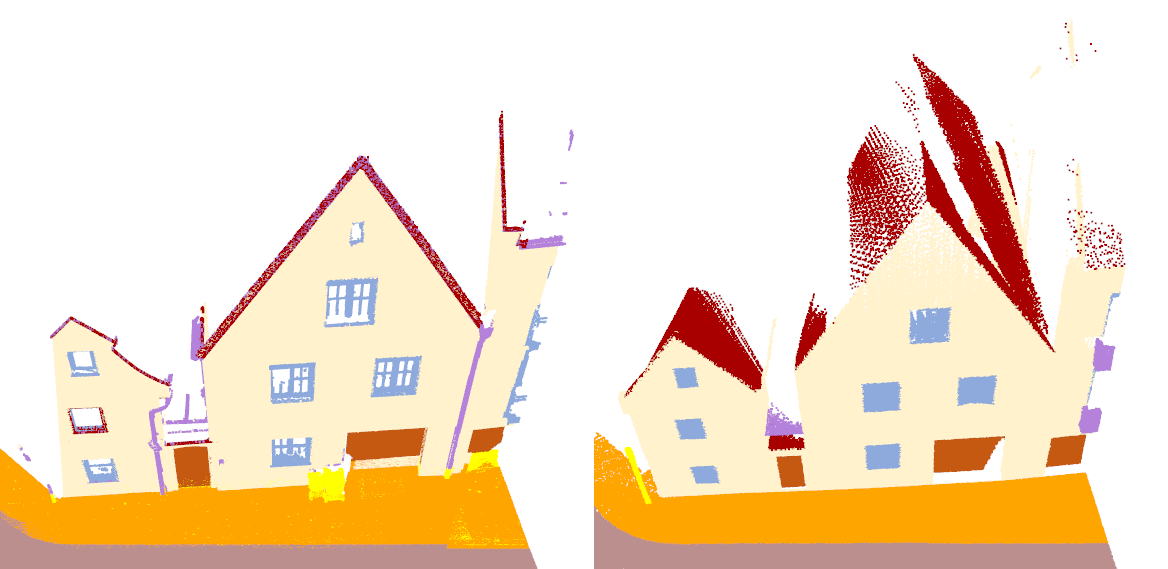}
    \caption{\texttt{RoofSurface} objects in real-world data (left) and the point cloud acquired from its synthetic twin (right).}
    \label{fig:roof}
\end{figure}

One of the fascinating results is the increase in model performance on the \texttt{RoofSurface} class when the ratio of synthetic data increases. 
From real-world point clouds, the \texttt{RoofSurface} objects are usually incomplete or even missing. 
They typically have very few points and only cover small parts of the entire roof. 
In synthetic data, with more flexibility in the control of driving route, angle, and height, the \texttt{RoofSurface} class data we acquired were more complete and covered relatively large areas of the roofs, as shown in~\autoref{fig:roof}. 
The \texttt{RoofSurface} class could serve as a compelling example of how synthetic data can enhance the performance of object segmentation tasks for rarely observed classes in real-world data.
By leveraging the high-level geometric structure of synthetic roof surfaces, the models achieved an $IoU^{(c)}$ of $0.65$ instead of only $0.01$ when using real-world data alone.

Overall, we observe that the synthetic point clouds simulated from semantic 3D city models provide two advantages to serve as sources for synthetic data: (1) The ability to preserve structures for objects that possess unique shapes, such as objects in classes \texttt{SolitaryVegetationObject}, \texttt{Window}, and \texttt{RoofSurface}. 
(2) For the parts missing or occluded in the real-world city point cloud, generating synthetic point clouds from their counterparts in the real-city digital models is a low-cost alternative.
On the other hand, some classes, lacking significant features, require refinement to provide unique and meaningful information for recognition. 
In some cases, such as objects geometrically represented by surfaces or objects in the class \texttt{Window}, the model could enhance their features by mixing a portion of the real-world data.
\section{Limitations}
Our method was primarily designed for the off-the-shelf laser scanner sensor parameters, enabling generic testing without the need to follow manufacturer-specific requirements.
This trait also may have influenced the domain gap experiments, as it inevitably introduced sensor-to-sensor parameter deviation.
Yet, our method can also be tested on any sensor-specific settings upon providing the manufacturer's details.

We have tested our method on the approximate driving trajectory of the actual \gls{MLS} measuring campaign, mimicking the right-hand traffic and keeping a center lane.
Consequently, the diverging laser-to-object distance may have occurred and impacted the laser scanning patterns and point cloud density.
Nevertheless, our approach allows for loading the actual driving trajectory, if available.

By using the to-date largest real-world-based and highly-detailed semantic road space models, including \gls{LOD}3 data, we ensured a high level of semantic and geometric information in our virtual testbed. 
Simultaneously, there were still instances that merely approximate the actual shape of objects, such as tree models (e.g., \autoref{fig:synthetic_tree}).
Also, dynamic objects, such as cars and pedestrians, were not present in the datasets.
Future work shall investigate creating even more advanced 3D reality-based virtual testbeds.

Even though we have leveraged a large dataset in the city of Ingolstadt with 100M points in our work, scalability remains a concern. 
This is because certain features from object classes, such as building facades or building installations, exhibit various characteristics that the neural network models struggle to capture with limited instances.

Furthermore, our evaluation methods employed two well-established semantic segmentation models trained on our datasets. 
However, the extended evaluation may also use the latest state-of-the-art models \citep{liu2024point,zeng2024self}, such as transformer-based approaches \citep{zhao2021point,wu2024point} and other datasets, that can complement the analysis.

Worth noting is that in our experimental approach, we sample points from the same geographical regions under the introduced splits.
Yet, the framework also allows for testing disjoint geographical analysis within the borders of the provided 3D model. 
Such an approach may open new application: If the point cloud splits are positively validated under the geographically disjoint areas of a small dataset subset (e.g., city district) and target ratios are established, the full-scale 3D model (e.g., whole city) can be used for simulation and only complemented by expensive manual annotations where required. 
Eventually, this leads to significant process automation and cost reduction.

\section{Conclusion and Outlook}
In this work, we present a method to measure the domain gap between simulated and real-world point clouds both deterministically and stochastically by leveraging digital replicas of real cities and real-world point clouds.
We also introduce a novel metric,  \emph{DoGSS-PCL}, measuring the quality of synthetically generated point clouds, which effectively measures both the semantic and geometric point cloud quality.
The introduced analysis shall serve as a blueprint for future works investigating domain gaps in simulated point cloud generation. 

Notably, the experiments also corroborate that our method generates simulated semantic point clouds that can complement real-world point clouds.
Intuitively, overall best performance is maintained when using only real and manually annotated point clouds.
Yet, we show that the 50\%-50\% fusion of real and simulated point clouds provides comparable segmentation results to solely using real point clouds, as there is only a 1\% overall accuracy difference.
Interestingly, we also observe that the ratios are non-linear and largely depend on the target class, e.g., drastic decrease (by 70\%) for \texttt{BuildingInstallation} when adding only even 25\% of synthetic point clouds, or slight increase for \texttt{Window} when adding 50\% of glass-rich synthetic point clouds.
By analyzing this trait, we also conclude that highly detailed semantic 3D city models can be used as semantic-rich training sets for developing data-demanding deep learning methods. 

Although the presented results are promising, testing sample size implies that caution must be exercised.
We plan to extend the acquired dataset for future work with additional 3D models and point clouds representing various architectural styles and traffic scenarios.
A comprehensive analysis of other simulation software shall also be undertaken, including HELIOS++, which comprises detailed sensor model specifications.
Another aim is to conduct a geographically disjoint analysis of the real-world 3D model to corroborate the method's usefulness as validation for large-scale applications, underscoring its potential to minimize labour-intensive manual annotations and leverage available 3D models. 


\section*{Declarations}


\paragraph{Acknowledgements}
We would like to thank Florian Hauck for setting up and designing the experiment for PointNet++. 
Furthermore, we would also like to thank the Munich Data Science Institute (MDSI) and Dr.\ Ricardo Acevedo Cabra for the TUM Data Innovation Lab, in the context of which the work was carried out.

\paragraph{Funding}
This work is supported by the German Federal Ministry of Transport and Digital Infrastructure (BMVI) within the \textit{Automated and Connected Driving} funding program under the Grant No. 01MM20012K (SAVeNoW).

\paragraph{Conflicts of interest/Competing interests}
All authors declare that they have no
competing interests.

\paragraph*{Availability of data and material}
The semantic models are available under \url{https://github.com/savenow/lod3-road-space-models}. 

\paragraph{Code availability}
\label{par:gitlab_repo}
The implementation of the methodology is available under \url{https://github.com/tum-gis/mind-the-domain-gap}. 

\paragraph{Authors' contributions}

Nguyen Duc: Methodology, Software, Validation, Formal analysis, Investigation, Writing - Original Draft.
Yan-Ling Lai: Methodology, Software, Validation, Investigation.
Patrick Madlindl: Methodology, Software, Validation, Writing - Original Draft.
Xinyuan Zhu: Methodology, Software, Validation.
Benedikt Schwab: Conceptualization, Writing - Original Draft, Writing - Review \& Editing, Visualization, Supervision.
Olaf Wysocki: Conceptualization, Writing - Original Draft, Writing - Review \& Editing, Supervision.
Ludwig Hoegner: Resources, Writing - Review \& Editing, Supervision.
Thomas H. Kolbe: Resources, Writing - Review \& Editing, Supervision.

\bibliographystyle{spbasic}      
\bibliography{bibliography}   

\end{document}